\documentclass[10pt,twocolumn,letterpaper]{article}

\usepackage{cvpr}
\usepackage{times}
\usepackage{epsfig}
\usepackage{graphicx}
\usepackage{amsmath}
\usepackage{amssymb}

\usepackage{times}
\usepackage{epsfig}
\usepackage{graphicx}
\usepackage{float}
\usepackage{wrapfig}

\usepackage{amsmath,amssymb}

\usepackage{bm,xspace}
\usepackage{comment}
\usepackage{verbatim}
\usepackage{multirow}
\usepackage{balance}
\usepackage{url}
\usepackage{booktabs}
\usepackage{etoolbox,siunitx}
\usepackage{calc}
\usepackage{pifont,hologo}
\usepackage{color}

\newcommand{\ra}[1]{\renewcommand{\arraystretch}{#1}}

\usepackage[super]{nth}
\usepackage{nicefrac}
\def\aa{\mathbf{a}}

\def\mm{\mathbf{m}}

\def\oo{\mathbf{o}}

\def\BB{\mathbf{B}}

\def\MM{\mathbf{M}}

\def\aA{\mathcal{A}}

\def\Re{\mathbb{R}}

\DeclareMathOperator*{\argmax}{arg\,max}

\DeclareMathSymbol{@}{\mathord}{letters}{"3B}

\newcommand\tuple[1]{\left\langle#1\right\rangle}

\definecolor{alexey}{rgb}{0.8,0.,0.8}

\newcommand\mypara[1]{\vspace{1mm}\noindent\textbf{#1}}

\def\latex/{\LaTeX}
\def\bibtex/{\hologo{BibTeX}}

\usepackage{blindtext}

\usepackage[pagebackref=true,breaklinks=true,letterpaper=true,colorlinks,bookmarks=false]{hyperref}

\cvprfinalcopy 

\def\cvprPaperID{86} 

\newcommand{\bdfpfull}{Belief DFP}
\newcommand{\bdfp}{BDFP}

\definecolor{dgreen}{RGB}{44,160,44}
\definecolor{dyellow}{RGB}{255,127,14}
\definecolor{dred}{RGB}{214,39,40}
\newcommand{\dgreen}[1]{\textcolor{dgreen}{#1}}
\newcommand{\dyellow}[1]{\textcolor{dyellow}{#1}}
\newcommand{\dred}[1]{\textcolor{dred}{#1}}

\newcommand{\yyes}{\dgreen{yes}}
\newcommand{\nno}{\dred{no}}
\newcommand{\apprx}{\dyellow{video}}

\ifcvprfinal\fi
\begin{document}

\title{Benchmarking Classic and Learned Navigation in Complex 3D Environments}

\author{Dmytro Mishkin\thanks{Work done during an internship at Intel Labs. Contact email address: \href{mailto:ducha.aiki@gmail.com}{\ttfamily ducha.aiki@gmail.com}. Project web page: \url{https://sites.google.com/view/classic-vs-learned-navigation}.}\\
Czech Technical University
\and
Alexey Dosovitskiy\\
Intel Labs
\and
Vladlen Koltun\\
Intel Labs
}

\maketitle

\begin{abstract}
Navigation research is attracting renewed interest with the advent of learning-based methods.
However, this new line of work is largely disconnected from well-established classic navigation approaches.
In this paper, we take a step towards coordinating these two directions of research.
We set up classic and learning-based navigation systems in common simulated environments and thoroughly evaluate them in indoor spaces of varying complexity, with access to different sensory modalities.
Additionally, we measure human performance in the same environments.
We find that a classic pipeline, when properly tuned, can perform very well in complex cluttered environments.
On the other hand, learned systems can operate more robustly with a limited sensor suite.
Both approaches are still far from human-level performance.
\end{abstract}


\section{Introduction}

Agile navigation in complex three-dimensional environments is a crucial capability for an intelligent agent operating in the physical world.
To achieve true autonomy, the agent should effectively and robustly navigate not only in areas that have been mapped in advance, but also when deployed in previously unseen environments.

Autonomous navigation has a long history in robotics and computer vision.
A classic approach is to design a modular pipeline that decomposes the problem into a sequence of sub-tasks, such as mapping, localization, planning, and low-level control.
Each of the sub-tasks can then be addressed separately with an appropriately engineered solution.
Methods of this type have been successfully deployed on a variety of mobile robotic platforms~\cite{Thrun2006,Bloesch2010,Hutter2016}.

Despite decades of research, existing robotic navigation algorithms are still less robust, generalizable, and scalable than navigation strategies employed by animals.
This motivates the recent surge of interest in applying learning techniques to navigation.
Learning-based approaches, in contrast with prior classic methods, do not require extensive hand-engineering, but rather promise to learn the full navigation system directly from data.
Potentially, a learning method can optimally exploit regularities in the data and therefore achieve performance superior to hand-crafted systems.
Indeed, deep convolutional networks trained end-to-end with reinforcement learning have demonstrated promising progress on a variety of navigation tasks~\cite{Mnih2016,Mirowski2017,Bruce2018}.

\begin{figure}
	\centering
	\setlength{\tabcolsep}{1mm}
	{
	\begin{tabular}{cc}
         \raisebox{-.5\height}{\includegraphics[width=0.45\linewidth]{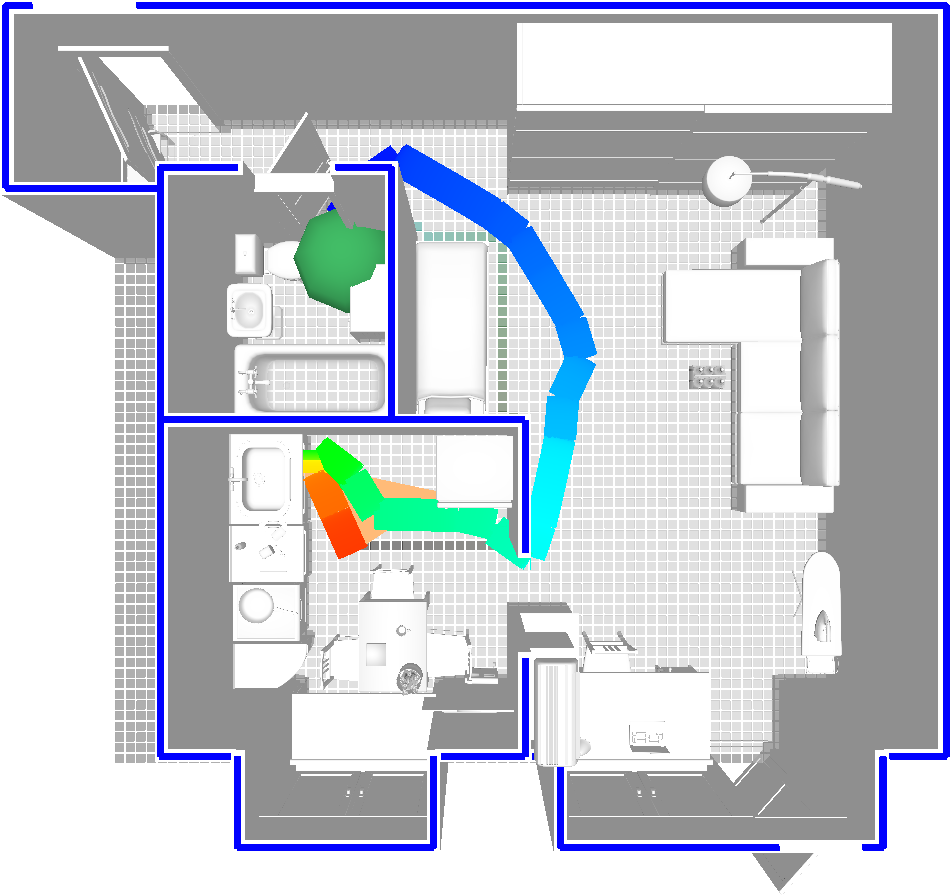}}  &
         \raisebox{-.5\height}{\includegraphics[width=0.45\linewidth]{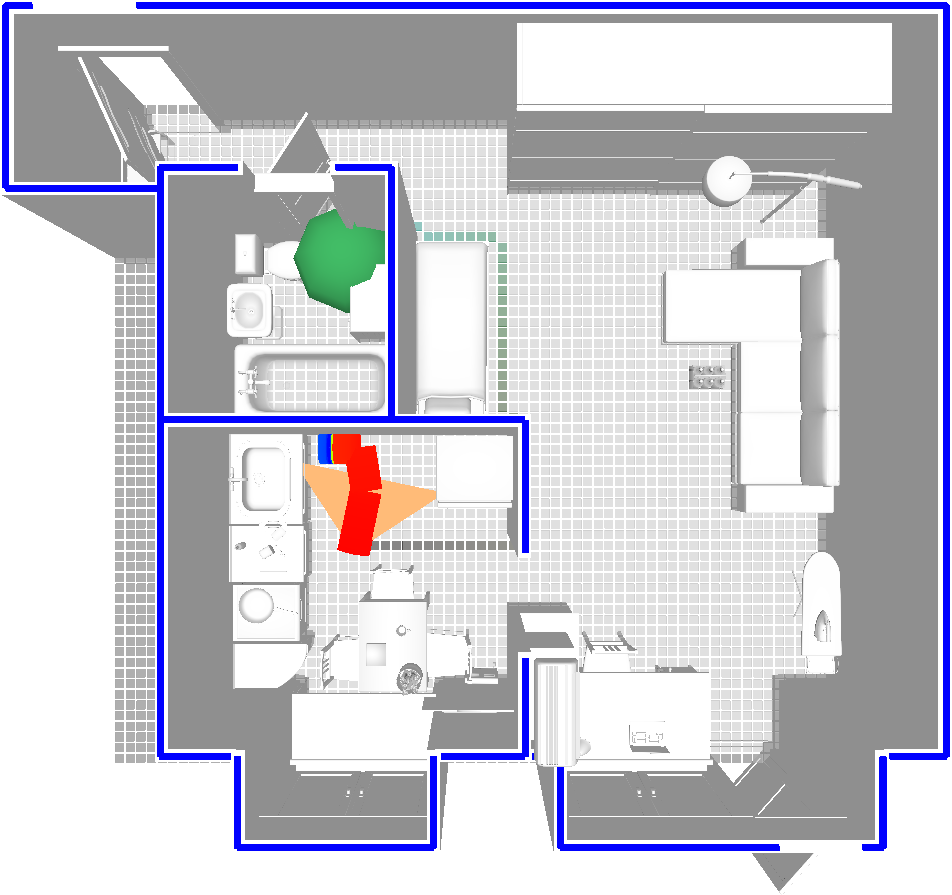}}
         \\
         \raisebox{-.5\height}{\includegraphics[width=0.45\linewidth]{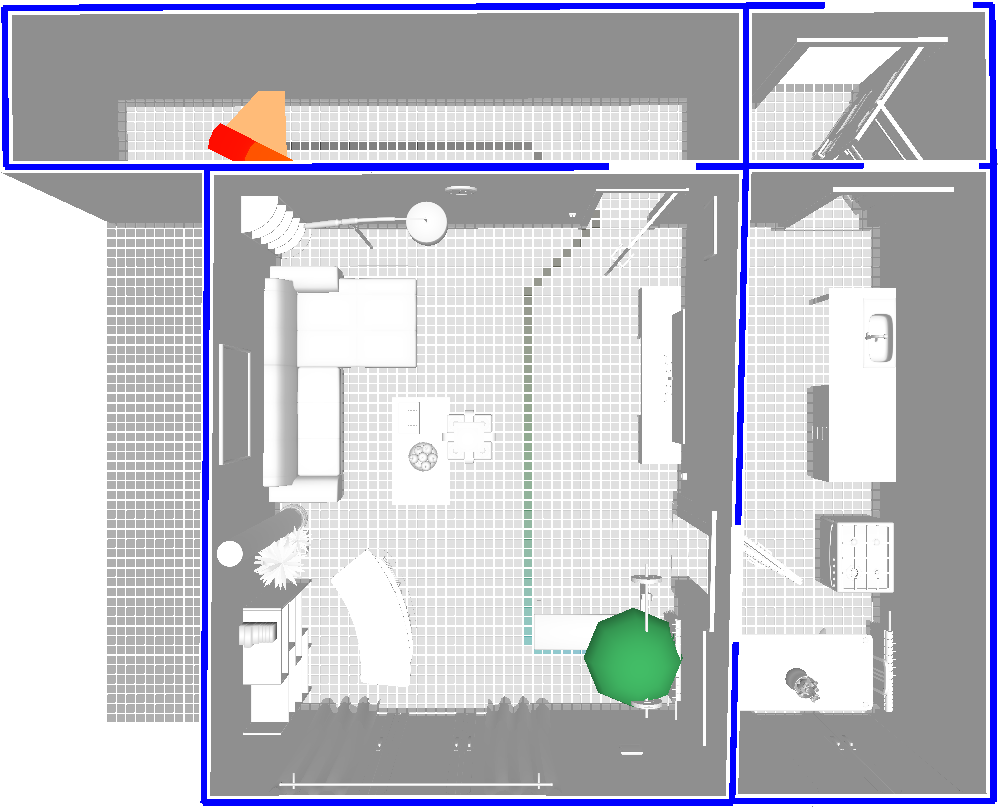}} &
         \raisebox{-.5\height}{\includegraphics[width=0.45\linewidth]{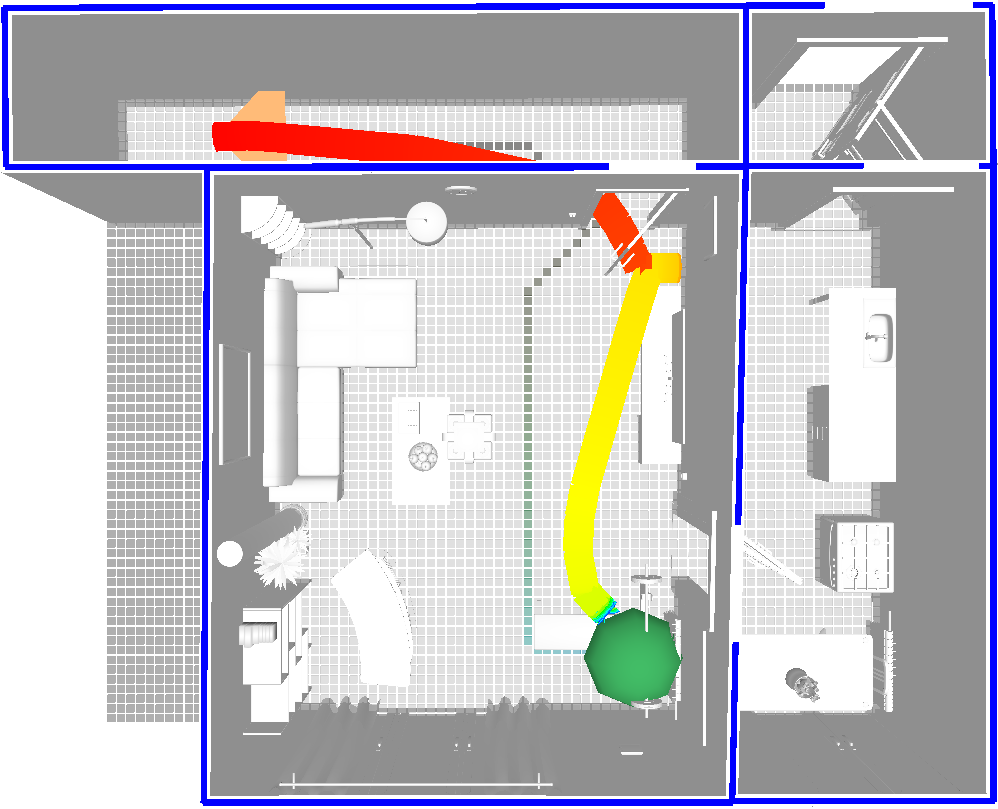}} \vspace{2mm}\\
          Classic pipeline & Learned agent
    \end{tabular}
    }
    \vspace{3mm}
	\caption{Success and failure cases of a classic navigation pipeline (left) and a learned agent (right). Starting position denoted by an orange triangle, the goal by a green sphere, the agent's path by a thick solid line, and the shortest path by a thin dotted line. Top: classic system successfully recognizes obstacles and navigates around them, while the learned agent tries to go straight for the goal ignoring the obstacles. Bottom: the localization module of the classic pipeline fails in a textureless narrow corridor, while the learned agent succeeds.}
	\label{fig:teaser}
	\vspace{-2mm}
\end{figure}


\begin{table*}
\centering
{\small
\label{tbl:related_work}
\begin{tabular}{lccccccc}
\toprule
                                                     & & Task         & No GT map       & No GT pose   & Continuous state & Locomotion   \\ \midrule
DPF, PF-net~\cite{Jonschkowski2018,Karkus2018}       & & Localization & \nno  & \yyes & \yyes & \nno  \\
ANL~\cite{Chaplot2018}                               & & Localization & \nno  & \yyes & \nno  & \yyes \\
MapNet~\cite{HenriquesVedaldi2018}                   & & Mapping      & \yyes & \yyes & \nno  & \nno  \\
EgoSM~\cite{Zhang2018}                               & & Mapping      & \yyes & \nno  & \yyes & \nno  \\
VIN, GPPN~\cite{Tamar2016,Lee2018}                   & & Planning     & \nno  & \nno  & \nno  & \yyes \\
QMDP-Net, MACN~\cite{Karkus2017,Khan2018}            & & Planning     & \nno  & \yyes & \nno  & \yyes \\
CogMap~\cite{Gupta2017}                              & & Navigation   & \yyes & \nno  & \nno  & \yyes \\
Neural Map~\cite{ParisottoSalakhutdinov2018}         & & Navigation   & \yyes & \nno  & \yyes & \yyes \\
SPTM~\cite{Savinov2018}                              & & Navigation   &\apprx & \yyes & \yyes & \yyes \\
General RL~\cite{Jaderberg2017,DosovitskiyKoltun2017}
                                                     & & Navigation   & \yyes & \yyes & \yyes & \yyes \\
This paper                                           & & Navigation   & \yyes & \yyes & \yyes & \yyes \\
\bottomrule
\end{tabular}
}
\vspace{2mm}
\caption{Learning-based approaches to mapping, localization, planning, and navigation. Most approaches are not applicable to autonomous navigation in previously unseen environments because they rely on unrealistic assumptions.}
\vspace{-2mm}
\end{table*}

Given these two classes of navigation approaches~-- classic modular pipelines and end-to-end learned models~-- a natural question arises: How do the two types of approaches compare to each other?
Does one dominate the other?
Or do both methodologies have characteristic strengths and weaknesses?
The answers to these questions are not known, since the two lines of research have been largely incompatible in terms of application areas and evaluation protocols.
Classic navigation approaches have been typically deployed on custom robotic platforms or in small-scale simulations, complicating extensive reproducible experimentation.
In contrast, new learning-based methods are usually evaluated in specialized simulated environments~\cite{Mnih2016,Savva2017,Gupta2017,HenriquesVedaldi2018,Xia2018}.
Due to these methodological differences, direct comparison has not been undertaken.

In this paper, we take a step towards coordinating and reconciling the two directions of navigation research by performing a fair and thorough comparison of classic modular and learned methods.
To this end, we set up approaches of both types in simulated indoor environments.
We use a basic navigation setup where the position of the goal relative to the agent is known at every time step.
We evaluate the methods on environments of different types (empty vs furnished, synthetic vs scanned) and with different sensory modalities provided to the agents (none, RGB, or RGB-D).

Our experiments show that the classic pipeline is generally strong and, when provided with RGB-D input, outperforms learned approaches by a large margin in cluttered and visually rich environments (Figure~\ref{fig:teaser}, top).
The learning-based agent, on the other hand, is more robust when the sensor suite is restricted to color images and in poorly textured areas (Figure~\ref{fig:teaser}, bottom).
Still, approaches of both types underperform compared to a human subject navigating in the same environments.
Given the complementary strengths of classic and learned approaches, we hypothesize that hybrid systems combining the best of both worlds may be most successful in the future.

\section{Related Work} \label{sec:related_work}

Navigation research has a long history in robotics~\cite{Bonin-Font2008,Thrun2005}.
A typical navigation pipeline consists of several modules responsible for sub-tasks of the navigation problem: mapping the environment, localizing the agent against the map, planning a path given the map, and following this path.
Each sub-task has been studied extensively.
Mapping and localization are typically treated jointly as the simultaneous localization and mapping (SLAM) problem.
Most SLAM methods are based on metric maps~\cite{Durrant-WhyteBailey2006,Cadena2016}, although topological methods have been developed as well~\cite{Garcia-FidalgoOrtiz2015}.
A survey of classic planning methods is provided by LaValle~\cite{LaValle2006}.
Classic navigation pipelines have been deployed on a variety of robotic platforms, including wheeled~\cite{Thrun2006}, legged~\cite{Hutter2016}, and aerial~\cite{Bloesch2010}.


The navigation problem has also been addressed by learning-based approaches.
Navigation can be interpreted within the framework of reinforcement learning (RL) and addressed with general-purpose RL algorithms~\cite{Mnih2016, Mirowski2017, Jaderberg2017, DosovitskiyKoltun2017, Zhu2017, Savva2017, Espeholt2018, Wu2018, Bruce2018}.
Extending RL agents with specialized memory architectures allows for implicit mapping~\cite{Oh2016, ParisottoSalakhutdinov2018}.
While RL-based approaches have demonstrated impressive human-level navigation in maze-like environments~\cite{Espeholt2018}, they may require a prohibitive number of training samples to reach this performance (up to billions~\cite{Espeholt2018}) and have only shown limited success when deployed in previously unseen realistic large cluttered indoor environments~\cite{Savva2017,Wu2018}.

In this line of research, most related to our work is the technical report of Bhatti~\etal~\cite{Bhatti2016}.
The authors experiment with providing the output of a SLAM algorithm to an RL agent operating in a simulated 3D environment.
However, the evaluation is constrained to a single task in a single small unrealistic environment.
In contrast, we conduct experiments in multiple large realistic cluttered environments with a proper train/validation/test split.
Moreover, we do not only use a SLAM algorithm, but a full navigation pipeline including path planning and locomotion.


\begin{figure*}
	\centering
	{\small
	\begin{tabular}{c}
        \includegraphics[width=0.8\linewidth]{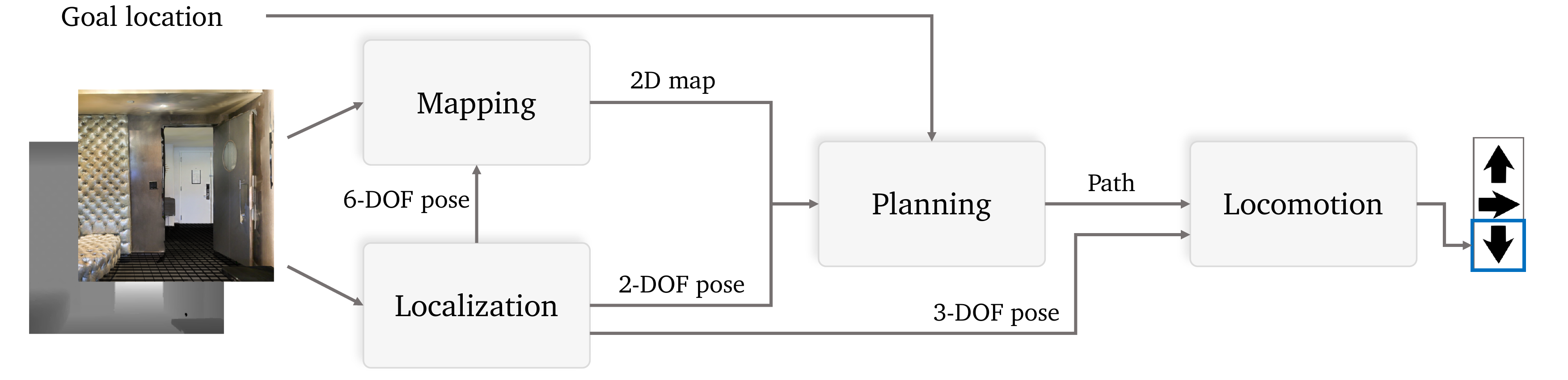}
    \end{tabular}
    }
    \vspace{2mm}
	\caption{Our modular navigation pipeline. Navigation is decomposed into 4 sub-tasks: mapping, localization, planning, and locomotion. The mapping module outputs a two-dimensional obstacle map of the environment. The localization module estimates the agent's pose. Together, these are used to plan a path to the goal. Given the planned path, the locomotion module selects an action to follow the path.}
	\vspace{-2mm}
	\label{fig:modular}
\end{figure*}

An alternative to using general RL architectures is to accept the classic decomposition of the navigation problem and learn specialized modules for localization, mapping, and planning.
For mapping, learning of both metric~\cite{Gupta2017,HenriquesVedaldi2018,Zhang2018} and topological~\cite{Savinov2018} maps has been explored.
Localization has been addressed with methods based on Bayes filters~\cite{Chaplot2018}, including particle filters~ \cite{Jonschkowski2018,Karkus2018}.
Finally, learning for planning has been studied both in the fully observable scenario~\cite{Tamar2016,Lee2018} and under partial observability~\cite{Karkus2017,Khan2018}.
For each task, learning-based approaches have been shown to outperform their hand-crafted counterparts in certain conditions.

However, as summarized in Table~\ref{tbl:related_work}, each of these works makes unrealistic assumptions about the discrete nature of the state space, availability of the ground-truth map or the agent's pose, or prior experience with the environment.
Therefore, to our knowledge, none of these methods can currently address the realistic autonomous navigation problem we are studying: reaching goals in a previously unseen environment in a continuous state space and without any ground truth information.
We expect future research to overcome these limitations and hope that our work will stimulate the development of practical learning-based navigation pipelines.

\section{Methods} \label{sec:method}

\subsection{Classic modular navigation pipeline}
Modular navigation pipelines have been widely used in robotics for decades, yet it is difficult to find a complete functional implementation that could be readily deployed in simulated environments.
We therefore implement our own solution, illustrated in Figure~\ref{fig:modular}.
Following standard practice, we decompose the navigation problem into a sequence of sub-tasks: localization, mapping, planning, and locomotion.
Each of these sub-tasks is solved by a separate module based on established algorithms.
We now describe each module.
Further details on all modules are provided in the supplement.

\begin{figure}
	\centering
    \includegraphics[width=\linewidth]{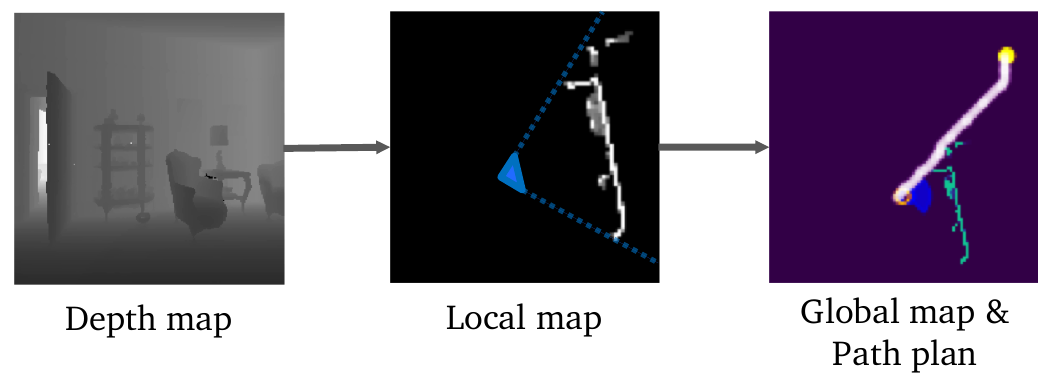}
	\caption{Illustration of the mapping and planning modules of the classic pipeline. A local map is generated by reprojecting the depth map to the top view and binning. Using the estimated agent's pose, the local map is integrated into the global map. The global map is used to plan a path towards the goal.}
	\vspace{-2mm}
	\label{fig:modular_illustration}
\end{figure}

\mypara{Localization.}
The objective of the localization module is to estimate the pose of the agent in the environment.
To achieve this, we use ORB-SLAM2, a popular real-time SLAM library~\cite{MurArtalTardos2017}.
We experiment with two variants of the method: accepting only a color image as input at each step (RGB) or accepting a color image and a depth map (RGB-D).
Given the inputs, the algorithm maintains a sparse keypoint-based map of the environment and at each step outputs an estimated 6-DOF metric camera pose.
The camera pose estimated by ORB-SLAM2 is the output of the localization module.
In case the algorithm fails (typically because tracking is lost due to difficulty of keypoint matching), it outputs a special ``Failure'' flag.
In such case there are two options~-- try to re-localize on the existing map or reset the map.
We found that in practice the latter option gives better results and use that in our experiments.
To resolve scale ambiguity in case of RGB-only input, we perform a hard-coded initialization procedure making use of the known height of the agent.


\mypara{Mapping.}
The mapping module estimates a two-dimensional obstacle map (Figure~\ref{fig:modular_illustration}, middle) that can be used by the planner to find a path towards the goal.
We use maps with a fixed cell size, 10x10 cm in our experiments.
We use two different mapping strategies depending on the availability of a depth sensor.

When depth maps are available, we re-project the observed points to the top view and discard points above the agent's height and on the ground level.
We then bin the points into map cells.
Each cell stores the maximum number of points ever binned into it.
Map cells that contain more points than a fixed threshold are marked as occupied.
An advantage of implementing mapping separately from SLAM is that the module could easily be extended with learnable components.

If depth map is not provided, we instead use the sparse 3D point cloud predicted by ORB-SLAM.
The rest is identical to the depth-based mapper, except that the threshold for a cell to be counted as an obstacle is set to a lower value.
The thresholds are tuned on the training set.

\mypara{Planning.}
Given the 2D obstacle map generated by the mapping module, the agent's pose (estimated by the localization module), and the goal location relative to the agent, we plan a trajectory to the goal (Figure~\ref{fig:modular_illustration}, right).
We pose path planning as finding the shortest path in the graph corresponding to the map: each traversable cell is a node, with \mbox{8-neighbor} connectivity and diagonal edges weighted by $\sqrt{2}$.
Since the environment is partially observable, the trajectory needs to be updated every time the map changes (for the purposes of planning, non-observed cells are considered traversable).
We employ the D* Lite algorithm~\cite{KoenigLikhachev2002} that is specifically designed to perform efficiently in this dynamic scenario.
When planning for the first time, it is identical to the A* algorithm.
For subsequent re-planning, D* Lite only updates the costs of the nodes that are affected by newly discovered obstacles.

\mypara{Locomotion.}
Finally, the locomotion module takes as input the planned path and generates an action to follow this path.
A standard solution would use a proportional-integral-derivative (PID) controller.
However, a PID controller normally requires a continuous action space, whereas in environments we are experimenting with the action space is severely discretized and consists of only three commands: move forward, turn left, and turn right.
We therefore design a simple custom rule-based controller inspired by PID.
We first select a waypoint on the planned path at a fixed distance $d_1$ meters from the agent.
If the agent's orientation is not within $\phi$ degrees from the direction to the waypoint, the agent rotates towards the waypoint.
Otherwise, the agent goes forward.
If the distance to the waypoint is below $d_2$ meters, a new waypoint is selected.
If the planned path changes, the waypoint position is updated accordingly.
Parameters $d_1$, $d_2$, and $\phi$ are tuned on the training set.

\subsection{End-to-end learned agent}
In this work we use Direct Future Prediction (DFP)~\cite{DosovitskiyKoltun2017} as the base end-to-end navigation algorithm, since this method is easy to implement and has been shown to perform well on navigation in 3D environments~\cite{DosovitskiyKoltun2017,Savva2017}, even when compared to more complex algorithms such as UNREAL~\cite{Jaderberg2017}.
The key idea of DFP is to learn sensorimotor control by predicting future values of behaviorally relevant quantities~-- measurements~-- via supervised learning.
We now briefly describe the algorithm and refer the reader to Dosovitskiy \& Koltun~\cite{DosovitskiyKoltun2017} for further details.

Consider an agent interacting with an environment in discrete time.
At each time step $t$ the agent receives from the environment an observation $\oo_t = \tuple{\mathbf{s}_t,\, \mm_t}$ that includes (typically high-dimensional) raw sensory information $\mathbf{s}_t \in \Re^{d_s}$ and a (typically low-dimensional) vector of measurements $\mm_t \in \Re^{d_m}$.
The key assumption of DFP is that the agent's objective function can be formulated as a function of future measurements ${u_t = U(\mm_{>t}),\;} {\mm_{>t} = \tuple{\mm_{t+1},\, \mm_{t+2},\, \ldots}}$.
The idea of the algorithm is to use a parametric function approximator $F(\cdot,\, \boldsymbol{\theta})$ with parameters $\boldsymbol{\theta}$ to estimate the expected future measurements for each possible action $\aa$ from a discrete action set $\aA \subset \Re^{d_a}$\,:
\begin{equation}
    \hat{\mm}_{>t}^\aa = F(\oo_{\leq t}, \aa, \boldsymbol{\theta}).
\end{equation}
Action selection then amounts to evaluation of the objective function on each of the predictions and choosing the optimal one:
\begin{equation}
    \aa_t = \argmax_{\aa \in \aA} \;U(F(\oo_{\leq t}, \aa, \boldsymbol{\theta})).
\end{equation}

In practice, we predict future measurements only for several future time steps, and output predictions for all possible actions at once by a single deep network.
We predict the distance and the direction to the goal, with the direction represented by its sine and cosine.


\paragraph{Interpretable learning with \bdfpfull{}.}
DFP can learn navigation in 3D environments~\cite{DosovitskiyKoltun2017},
however, the resulting policy is black-box and therefore difficult to interpret.
Since we are interested in better understanding the behavior of different types of navigation policies,
we designed \bdfpfull{} (\bdfp{})~-- a variant of DFP that is tailored to the navigation task and
provides additional insight into the inner workings of the policy.
The idea is to introduce an intermediate map-like representation in future measurement prediction.
The architecture is inspired by deep networks with attention,
as well as the successor representation~\cite{Dayan1993} and successor features~\cite{Barreto2017} in RL.
Further details and illustrations are provided in the supplement.

\section{Experiments} \label{sec:experiments}

\subsection{Experimental setup}

\paragraph{Environment.}
We use the MINOS simulator~\cite{Savva2017} as our evaluation platform.
We experiment with environments of two types.
The first type are synthetic house models from the SunCG dataset~\cite{Song2016}.
Environments are relatively simplistic, but the size and the diversity of the dataset allow for reliable experimentation.
We use two variants of SunCG: with furniture (``Furnished'') and with all furniture removed (``Empty'').
The second type of environments are 3D reconstructions of real indoor spaces from the Matterport3D dataset~\cite{Chang2017}.
This type of data is more realistic both in terms of visual quality and the environment layout and clutter, but the dataset is relatively small~-- 90 environments in total.

The agent in MINOS is represented by a cylinder with radius of 0.1~m and height 1.09~m.
The environments are flat, therefore the agent's pose can be represented by its 2D position and orientation. Both are continuous.
The action space is discrete and consists of 3 actions~-- go forward, turn left, turn right~-- which apply, respectively, linear or angular force to the agent.
Sensory input includes rendered RGB frames and depth maps at resolution 256x256 pixels for the classic agent and 128x128 for the learned ones.
ORB-SLAM does not detect enough keypoints on smaller images, while RL agents do not benefit from higher resolution.
The depth maps provide accurate depth information in the range of [0.001, 4.0] meters.

We use the train/validation/test split of environments from the MINOS repository, but we filter out the episodes for which the simulator is unable to compute the shortest path to the goal, as the shortest path is needed to compute some of our evaluation metrics.
During training, start and goal positions are randomly generated.
In SunCG, texture randomization is applied.
For validation and test we use fixed sets of start-goal pairs for each map.
We evaluate for 100 episodes on Matterport3D (20 start-goal pairs per map) and for 340 episodes on each of SunCG variants (10 start-goal pairs per map).

\begin{table*}
\centering
{\small
\begin{tabular}{lccccccccc}
\toprule
                        && \multicolumn{2}{c}{SunCG Empty} && \multicolumn{2}{c}{SunCG Furnished} && \multicolumn{2}{c}{Matterport3D}  \\
                        && Small    & Medium         && Small    & Medium   && Small  & Medium  \\ \midrule
UNREAL~\cite{Savva2017} && 72.9     & 63.2           && 64.1          & 45.3          && 38.0   & 20.0\textsuperscript{*}     \\
DFP~\cite{Savva2017}    && \textbf{80.3}     & \textbf{64.1}  && \textbf{64.5} & 43.6          && 27.3   & 18.2\textsuperscript{*}     \\
Ours-DFP                && 74.5     &  52.0          && \textbf{64.5} & 46.7          &&  -      & \textbf{41.0}\textsuperscript{*}  \\
Ours-\bdfp{}            && 78.1 & 55.0      &&  60.0         & \textbf{47.9} && -       & \textbf{41.0}\textsuperscript{*}   \\
\bottomrule
\end{tabular}
}
\vspace{2mm}
\caption{Comparison of learned navigation approaches in MINOS. We report average success rate. Our DFP implementation generally performs on par with previously reported results. \bdfpfull{} slightly outperforms DFP in most environments. \textsuperscript{*}The exact set of environments and start-goal pairs used for evaluation on Matterport3D is not identical, which may account for part of the performance difference.}
\vspace{-2mm}
\label{tbl:results_learned}
\end{table*}

\mypara{Task.}
We address goal-directed navigation with known position of the goal relative to the agent (PointGoal, according to the terminology of Anderson~\etal~\cite{Anderson2018}).
We chose this simplest scenario over alternative goal specification (for instance, by object class or by room type) to ensure that our evaluation focuses specifically on the navigation capabilities of agents, not semantic understanding of the environment.
The PointGoal task is far from trivial: without any prior knowledge of the environment, the agent needs to find the shortest possible path to the goal, avoid obstacles, and backtrack when facing a dead end.

The task is set up as a sequence of goal-directed navigation episodes.
In each episode the agent has to reach a fixed goal position from its initial location, given a limited time budget (50 seconds in our experiments, which corresponds to 500 actions).
The episode ends after the time budget has elapsed or when the agent takes the special ``Done'' action.
Since the distance to the goal is provided to all agents, we use a simple threshold on this value to output the ``Done'' action.
The agent does not carry any memory across episodes. That is, in the beginning of each episode it has no prior knowledge of the environment and has to explore the environment while navigating towards the goal.
The goal position relative to the agent is provided as a two-dimensional vector in the agent's egocentric coordinate system (in practice, we represent the vector in polar coordinates).
Note that this vector is insufficient to globally localize the agent relative to the goal: the agent has three degrees of freedom ($2$ for the position and $1$ for the orientation), while the vector to the goal only provides two values.

We use three evaluation metrics in our experiments: success rate (SR), success weighted by path length (SPL)~\cite{Anderson2018}, and ``pace''.
Success rate is simply the fraction of the episodes in which the agent reaches the goal within the time budget.
SPL factors in the length of the path taken by the agent: each successful episode contributes to the score proportionally to the ratio of the shortest path to the executed path.
Finally, pace considers time taken to reach the goal: it is the average fraction of the time budget the agent has left at the end of an episode.
Detailed definitions are provided in the supplement.
In what follows we report all metrics in \%.
Value of each of the metrics is between 0\% and 100\% and for all of them higher values are better.

\mypara{Model details.}
All learned modules are implemented by deep networks using the PyTorch framework~\cite{Paszke2017}.
Potential inputs to the models are: color image, depth map, actions from previous steps, distance to the goal, and the direction to the goal.
For each input, we feed a stack of 4 most recent observations to the agent.
Color and depth inputs are processed by convolutional networks, while actions and goal information are processed by fully connected networks.
The outputs of these feature extractor networks are concatenated and fed into a two-stream fully connected action-expectation network (see Dosovitskiy\&Koltun~\cite{DosovitskiyKoltun2017}).
This network outputs the future measurement predictions for all actions and future time steps at once.
For \bdfp{}, the expected position maps have resolution $25 \times 25$ and are predicted with fully connected layers.

We train the models with the Adam~\cite{Kingma2014} optimizer with learning rate 1e-5, standard momentum, and mini-batch size 32.
We train for a total of 5 million environment steps, corresponding to 5.8 days of real-time experience.
Learning rate is divided by 3 every 2M steps.
Four simulators are run in parallel.
During training we execute an $\epsilon$-greedy policy, starting with the probability of random action $\epsilon=0.95$, linearly decayed to $\epsilon=0.1$ at the end of the training.
We have experimented with other training setups (longer training runs of 8-10M samples, different learning rates, exploration schedules, network architectures), but have not observed significant performance improvements.
Further training details are provided in the supplement.

\begin{table*}
\centering
{\small
\begin{tabular}{lcccccccccccc}
\toprule
& & \multicolumn{2}{c}{SunCG Empty} & & \multicolumn{2}{c}{SunCG Furnished} & & \multicolumn{2}{c}{Matterport3D} & & \multicolumn{2}{c}{Mean} \\
& &  Classic & Learned & & Classic & Learned & & Classic & Learned & & Classic & Learned \\ \midrule
Blind  &&  52.3 & \textbf{52.9} &&  \textbf{44.4} & 13.8 && \textbf{40.9} & 27.7 && \textbf{45.9} & 31.4   \\
RGB    &&  9.4 & \textbf{49.9} &&  4.4 & \textbf{44.8} && 4.7 & \textbf{37.4} && 6.1 & \textbf{44.0}   \\
RGB-D  &&  \textbf{65.7} & 54.6 &&  \textbf{67.7} & 47.2 && \textbf{70.2} & 45.5 && \textbf{67.9} & 49.1  \\
\midrule
RGB-D + MINOS pose        && 82.5 & -- && 84.5 & -- && 78.8 & -- && 82.0 & --  \\
RGB-D + MINOS pose \& map && 81.8 & -- && 81.5 & -- && 85.1 & -- && 82.8 & --   \\
\bottomrule
\end{tabular}
}
\vspace{2mm}
\caption{Performance of a classic modular pipeline and a learned agent when provided with different input modalities. We report Success weighted by Path Length (SPL), as proposed by Anderson~\etal~\cite{Anderson2018}. For each environment and sensory input, the best of the two results is highlighted in bold. Classic pipeline performs well when provided with RGB-D input, but fails with RGB only.}
\vspace{-2mm}
\label{tbl:results_modalities}
\end{table*}

\subsection{Results}

\mypara{Comparison to prior learning methods.}
We start by comparing our implementation of learned agents with prior results reported in the literature.
We restrict the comparison to general RL algorithms, since, as explained in Section~\ref{sec:related_work}, none of the existing navigation-specific learning approaches can be deployed in our realistic navigation scenario.
Table~\ref{tbl:results_learned} shows that our DFP implementation generally performs on par with the results reported by Savva et al.~\cite{Savva2017}, and \bdfpfull{} outperforms standard DFP in all cases except one.
Motivated by these results, in what follows we use \bdfp{} as the main learning method.



\begin{figure}
	\centering
  \setlength{\tabcolsep}{0.5mm}
	{\small
	\begin{tabular}{cccc}
      \rotatebox[origin=c]{90}{Classic pipeline} &
      \raisebox{-.5\height}{\includegraphics[height=0.6\linewidth]{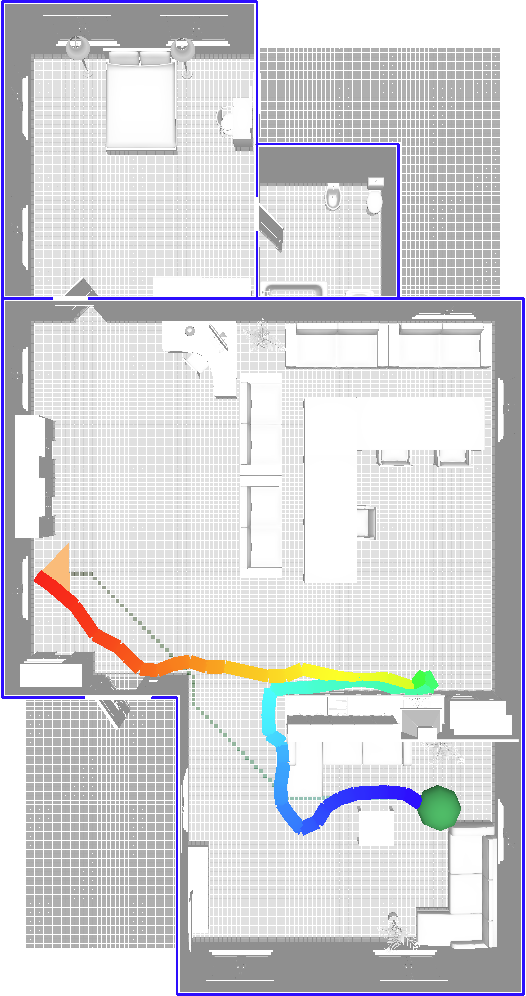}} &
      \raisebox{-.5\height}{\includegraphics[height=0.6\linewidth]{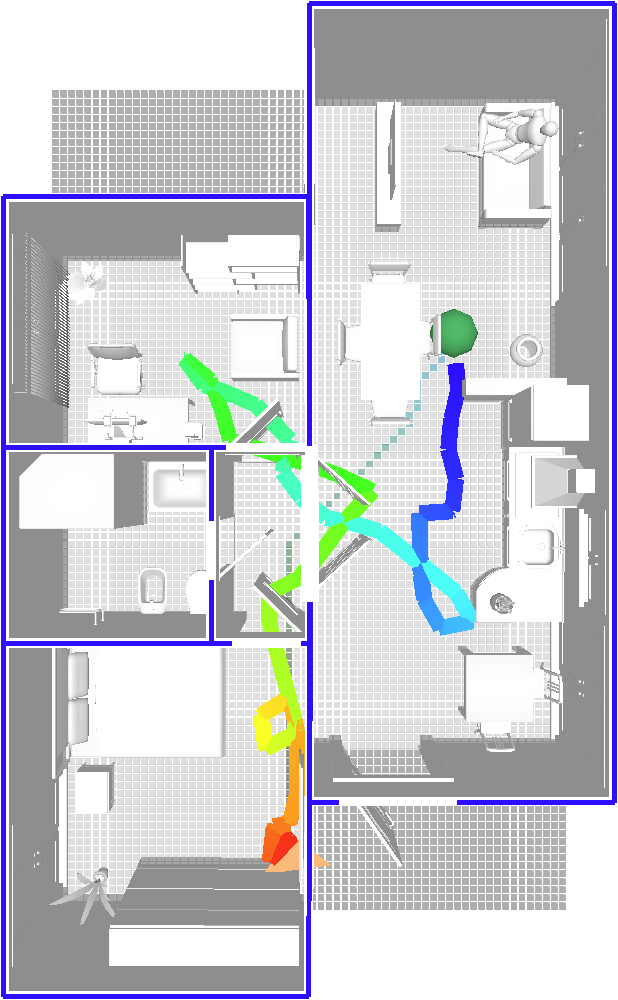}} &
      \raisebox{-.5\height}{\includegraphics[height=0.6\linewidth]{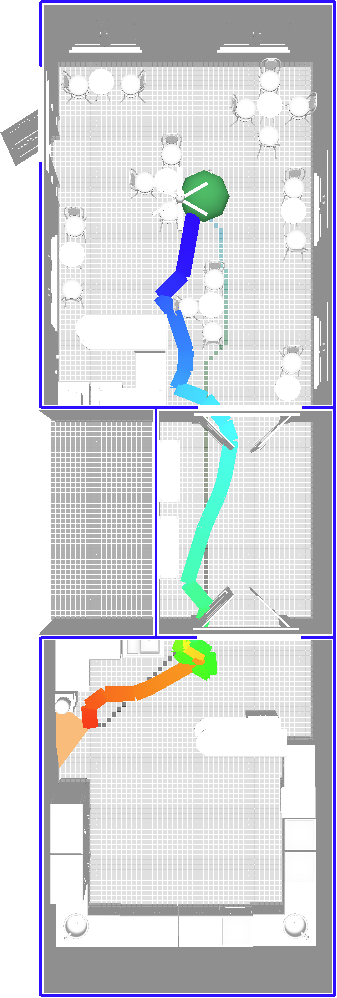}} \\
      \rotatebox[origin=c]{90}{Learned agent} &
      \raisebox{-.5\height}{\includegraphics[height=0.6\linewidth]{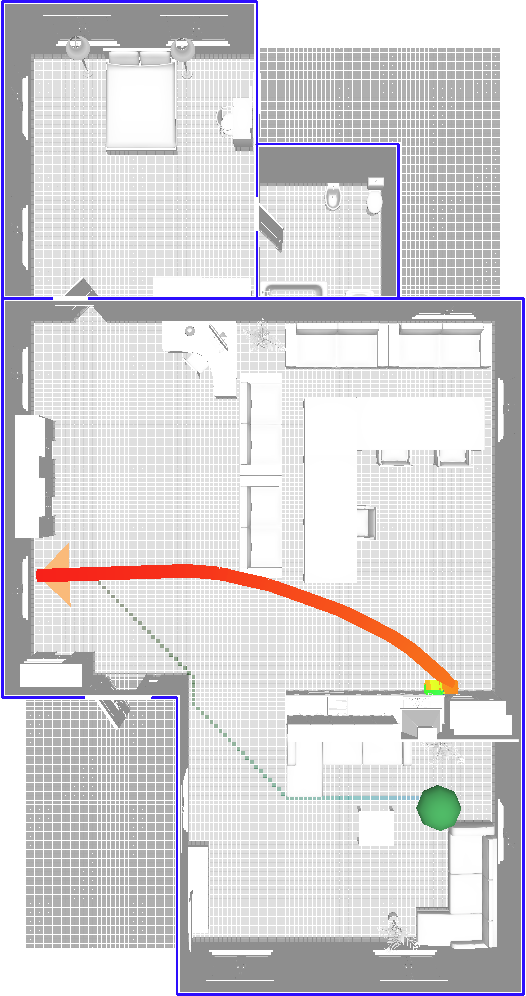}} &
      \raisebox{-.5\height}{\includegraphics[height=0.6\linewidth]{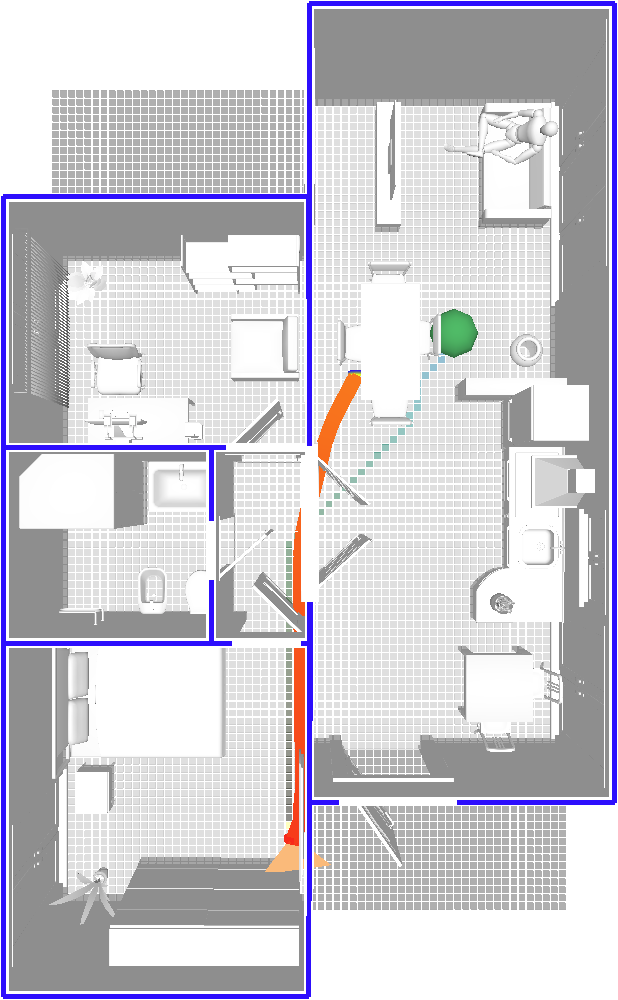}} &
      \raisebox{-.5\height}{\includegraphics[height=0.6\linewidth]{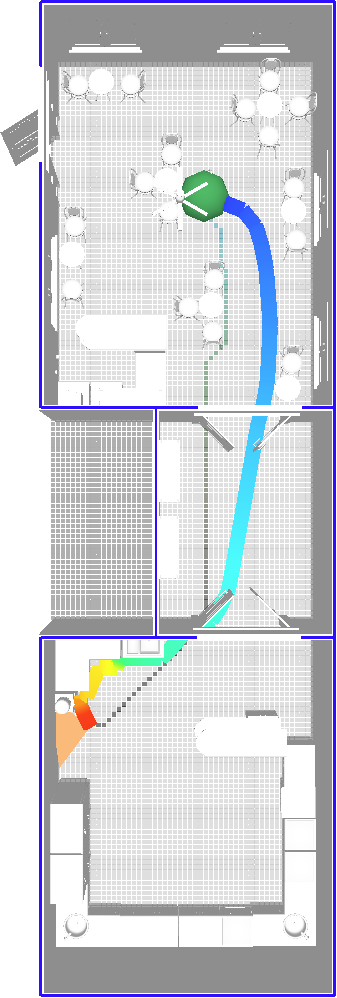}}  \\

    \end{tabular}
    }
    \vspace{2mm}
	\caption{Visualization of the trajectories of classic (top) and learned (bottom) agents in three navigation episodes. Trajectories of the classic agent are far from optimal, but do reach the goal in the end. The learned agent fails to navigate around obstacles. The supplementary video shows additional qualitative results.}
	\vspace{-4mm}
	\label{fig:navigation_qualitative}
\end{figure}

\mypara{Classic vs learned navigation.}
We now turn to the main subject of this work: comparing classic and learned algorithms.
We evaluate the modular pipeline and \bdfp{} on three environment types of increasing difficulty: SunCG Empty, SunCG Furnished, and Matterport3D.
Moreover, we experiment with providing different modalities as input to the agents: ``blind'' agents only get the direction and distance to the goal, ``RGB'' also get color images, and ``RGB-D'' additionally get depth maps.
Furthermore, to estimate the potential for improvement, we evaluate two more setups with extra information provided to the agent: agent's pose or both agent's pose and a map of the environment.
The resulting SPL scores are reported in Table~\ref{tbl:results_modalities} and other metrics are provided in the supplement.
These results lead to several interesting conclusions.

First, the ``blind'' classic agent is surprisingly successful, achieving SPL of 40-50\% in all environments.
This algorithm simply follows the straight path towards the goal, and the degree of its success suggests that many of the goals are reachable with this naive strategy.
We believe this agent should serve as a basic baseline for the PointGoal task, not a randomly acting agent as evaluated by Savva et al.~\cite{Savva2017}.

Second, the classic pipeline performs well with \mbox{RGB-D} input, but fails with RGB only.
The reason is the performance of the SLAM system.
When only provided with color images, the algorithm struggles under our conditions, which feature predominantly purely rotational or purely forward motion, often in presence of poorly textured surfaces.
As a result, instead of moving towards the goal, the agent spends time repeatedly re-initializing SLAM (including the time-consuming procedure for scale estimation).
The addition of depth input dramatically robustifies the method, allowing for reliable localization.

It may seem counter-intuitive that the performance of the RGB agent is much worse than that of the blind agent.
Indeed, we could set up the RGB agent to ignore the image and navigate directly towards the goal, in which case it would be identical to the blind agent.
However, this result would not be informative: we already report the performance of the blind agent anyway.
Instead, we try to actually make use of the color images and find that the performance of RGB SLAM is so poor that the resulting agent fails almost completely.
This is an informative finding that can inspire future work on improving RGB SLAM.

Third, the performance of the classic method improves for more complex cluttered environments.
The explanation is that the more objects are in the environment, the easier it is for the SLAM algorithm to find reliable keypoints and localize reliably.
Still, providing ground truth pose to the classic method brings substantial (8-16\%) further improvement compared to RGB-D input.
Provided map adds another 6\% SPL in Matterport3D environments.
These numbers suggest that the localization and mapping modules can still be improved.
Surprisingly, in SunCG environments planning with the provided map leads to worse performance than using the mapping module.
We believe this is caused by imperfections in the maps provided by MINOS (for instance, the maps do not include some doors that block entryways).

Finally, the learned approach only marginally outperforms the ``blind'' classic method and always performs much worse than the classic pipeline with RGB-D input.
This poor performance may be explained by the lack of an appropriate memory structure capable of maintaining an internal representation of the environment.
Another hypothesis is that reinforcement learning methods may require longer training time to reach optimal performance: we train for 5 million time steps (corresponding to 5.8 days of real-time experience), while some state-of-the-art RL systems take up to a billion time steps (corresponding to years of real-time experience)~\cite{Espeholt2018}.
We hope these results will stimulate the development of performant and sample-efficient learned navigation algorithms.

Figure~\ref{fig:navigation_qualitative} shows qualitative navigation results.
We plot trajectories executed by the agents, as well as the shortest path to the goal.
Trajectories of the classic method are not smooth and are often far from the shortest path, but the agent eventually reaches the goal by avoiding obstacles and backtracking from dead ends.
In contrast, the trajectories of the learned agent are smooth, but the agent often gets stuck near obstacles and is typically unable to make progress towards the goal after that.
We encourage the reader to watch the supplementary video for additional qualitative results. 

\begin{table}
\centering
{\small
\begin{tabular}{lccc}
\toprule
SLAM input & SunCG-E & SunCG-F & M3D  \\ \midrule
RGB   &  9.4  &  4.4  & 4.7    \\
RGB + MonoDepth~\cite{Hu2019} &  43.8 & 47.3 & 37.5   \\
RGB + StereoDepth~\cite{OpenCV2000} &  30.2  &  34.7 & 47.3   \\
RGB + GT Depth  &  65.7 &  67.7  & 70.2 \\
\bottomrule
\end{tabular}
}
\vspace{2mm}
\caption{Performance of the classic modular pipeline with different inputs to SLAM:
monocular color image (RGB), RGB plus depth estimated from a single image by a deep network (RGB + MonoDepth),
and RGB with depth estimated from a stereo pair (RGB + StereoDepth).
We report SPL (higher is better) on SunCG Empty (E) and Furnished (F), as well as Matterport3D (M3D).
Navigation results substantially improve when using depth estimates, but are still far from the performance with ground truth depth.}
\vspace{-2mm}
\label{tbl:depth_estimation}
\end{table}

\mypara{Classic agent with depth estimation.}
Table~\ref{tbl:results_modalities} shows a dramatic difference in the performance of the classic agent depending on the input provided to the SLAM system: color images or color and depth.
Motivated by this, and aiming to improve the results of the RGB classic agent,
here we experiment with estimating depth from color images and feeding these estimates to the SLAM system.

We study two setups: monocular and stereo.
For monocular depth estimation, we use the deep model of Hu et al.~\cite{Hu2019}.
To ensure best performance, we fine-tune ResNet-50 pre-trained on the NYU dataset (provided by the authors) on data extracted from MINOS, separately on SunCG and Matterport3D.
We then use the corresponding model for each of the datasets.
For stereo depth estimation, we use the
OpenCV~\cite{OpenCV2000} function StereoSGBM with weighted least squares disparity filter (see further details in the supplement).

The results are shown in Table~\ref{tbl:depth_estimation}.
Both depth estimation methods substantially improve the performance of the navigation agent.
Stereo matching performs well on the richly textured Matterport3D dataset, but struggles on SunCG with homogeneous textures and repeated patterns.
In contrast, monodepth works better on the structurally simpler SunCG dataset.
Overall, the performance with estimated depth is comparable to that of the blind agent (Table~\ref{tbl:results_modalities}).
The main failure mode of both stereo and mono depth estimation is that small metric imprecisions in depth prediction lead to degradation of the map and the pose estimated by SLAM.
Another issue is with major failures in depth estimation, for instance in scenes with large weakly textured areas or repetitive structures (see qualitative depth map estimation results in the supplement).
These issues might be alleviated by more advanced depth estimation techniques or improved filtering and aggregation of the predicted depth maps.

\begin{table*}
\centering
{\small
\begin{tabular}{lcccccccccccc}
\toprule
                & & \multicolumn{3}{c}{SunCG Empty} & & \multicolumn{3}{c}{SunCG Furnished} & & \multicolumn{3}{c}{Matterport3D} \\
                & &  Classic & Learned & Human      & & Classic & Learned & Human           & & Classic & Learned & Human        \\ \midrule
SPL             & &  65.7 & 54.6&  \textbf{90.5} && 67.7 & 47.2 &  \textbf{87.9} && 70.2 & 45.5 & \textbf{86.7}    \\
Success rate    & &  70.2 & 56.4 & \textbf{97.0} && 75.0 & 48.2  & \textbf{100.0}  && 75.0  & 47.0 & \textbf{95.3}    \\
Pace            & & 57.8 & 47.9 & \textbf{84.0} &&  58.3 & 39.0& \textbf{82.8} && 56.2  & 36.6 & \textbf{72.6}     \\
\bottomrule
\end{tabular}
}
\vspace{2mm}
\caption{Comparison to human performance on different navigation performance metrics. For all metrics, higher is better and the best possible result is 100\%. Both algorithms get RGB-D input, while the human navigates using RGB images only. Human performance surpasses both algorithms in all environments and metrics. In the most complex environments (Matterport3D), the advantage over the classic pipeline is the smallest.}
\vspace{-2mm}
\label{tbl:results_metrics}
\end{table*}

\mypara{Human baseline and different metrics.}
We now compare the two artificial navigation systems with a human.
To this end, a human subject performed $43$ navigation trials in Matterport3D and $68$ trials each in SunCG Empty and Furnished.
We made the comparison as fair as possible: the human subject used the exact same action frequency and action space as the artificial systems, and has to navigate based on 128x128 pixel RGB images.
One remaining inevitable advantage of the human is that when evaluating several start-goal pairs in the same environment, there is a chance that the human remembers the environment layout from previous trials.
Based on self-report, we do not expect this to significantly affect the performance.

\begin{figure}
	\centering
    \includegraphics[width=\linewidth]{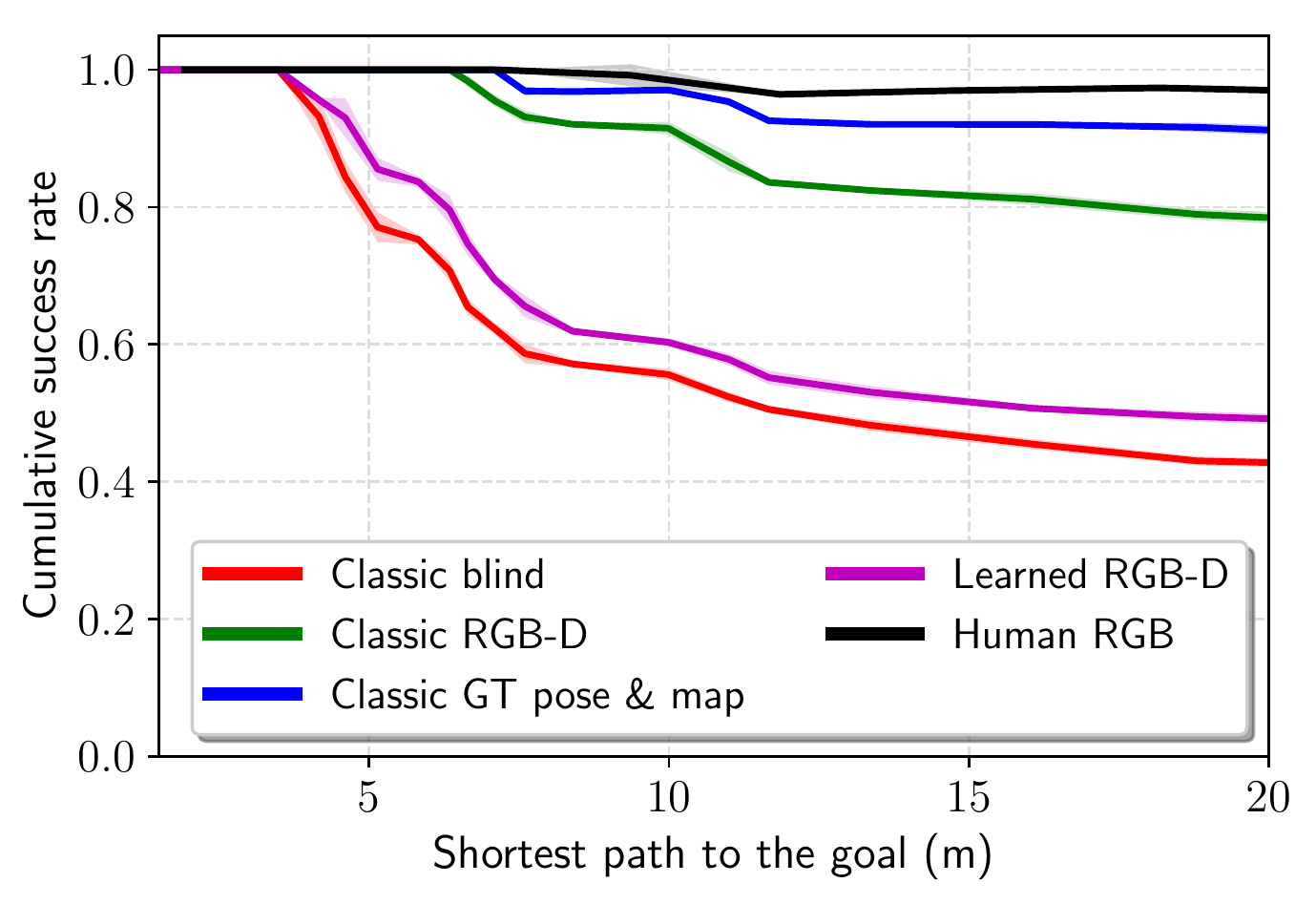}
	\caption{Performance of algorithms on Matterport3D as a function of the length of the shortest path to the goal. We plot cumulative success rate. That is, for a value $L$ on the $x$-axis we plot the average success rate in all episodes where the shortest path to the goal was shorter than $L$.}
  \vspace{-3mm}
	\label{fig:env_size}
\end{figure}

The results are summarized in Table~\ref{tbl:results_metrics}.
We evaluate three metrics: success rate, SPL, and pace.
Human performance exceeds that of algorithms in all environments and metrics, even though the human is navigating based on color images only, without depth.
Interestingly, in the most challenging Matterport3D environment the advantage is the smallest: SPL of 86.7\% versus 70.2\% for the classic pipeline.
Overall, all three metrics are correlated across different agents and environments.
In particular, the ranking of the results~-- learned, classic, human~-- is consistent for all metrics in all environments.

Figure~\ref{fig:env_size} shows the dependence of the success rate of different agents on the length of the shortest path to the goal in Matterport3D.
(Plots for other environments and metrics are shown in the supplement.)
As expected, the success rate of all methods drops for goals that are further away.
Human performance is most stable, exceeding even the classic agent with access to simulator-provided pose and map.
Learned RGB-D agent is only slightly better than the ``blind'' baseline.

Interestingly, human outperforms even the classic agent that has access to MINOS map and camera pose.
This is for two main reasons.
First, maps provided by MINOS are imperfect: for instance, some doors that are present in the environment are not in the maps.
Second, the locomotion controller, which converts the planned path to the actual motor commands, is not always able to perfectly track the optimal planned path.
This can be especially critical in confined spaces and narrow corridors.
The design of the locomotion controller is non-trivial, since the state space is continuous, while the action space is severely discretized.

\section{Conclusion} \label{sec:conclusion}
We have benchmarked a representative classic modular navigation pipeline against recent navigation approaches based on reinforcement learning and against a human operator. This evaluation led to several surprising findings.

First, the classic pipeline shows very strong performance when provided with RGB-D input.
It outperforms the learned agent by a large margin, especially in complex cluttered environments.
However, the classic method is sensitive to the set of modalities provided as input: without access to depth information it fails catastrophically.
The performance can be partially recovered by estimating the depth either from a single image or using a stereo setup.

Second, our learned agent generally performs poorly, only slightly better than a simple baseline that cuts directly to the goal.
This is not to say that learning is in principle incapable of good navigation, but further research is needed to make learning methods more robust and sample-efficient.
On the upside, unlike the classic system, the learning approach does not exhibit catastrophic failure when provided with RGB-only input.

Lastly, even given RGB-D input, ground truth agent's pose, and an approximate environment map provided by the simulator, both algorithms perform worse than a human navigating based only on low-resolution color images.

These results suggest that there is still significant room for improvement in artificial navigation algorithms.
We believe that a particularly promising direction of research is combining ideas from classic navigation with learning.



{\small
\bibliographystyle{ieee}
\bibliography{biblio}
}

\appendix

\section{Model Details}
\subsection{Learned agent}
The standard version of DFP can learn navigation in 3D environments~\cite{DosovitskiyKoltun2017}. However, the resulting policy is black-box and therefore difficult to interpret.
Since we are interested in better understanding the behavior of different types of navigation policies, we designed \bdfpfull{} (\bdfp{})~-- a variant of DFP that is tailored to the navigation task and provides additional insight into the inner workings of the policy.
The architecture is inspired by deep networks with attention, as well as the successor representation~\cite{Dayan1993} and successor features~\cite{Barreto2017} in RL.

\begin{figure*}
	\centering
    \includegraphics[width=0.8\linewidth]{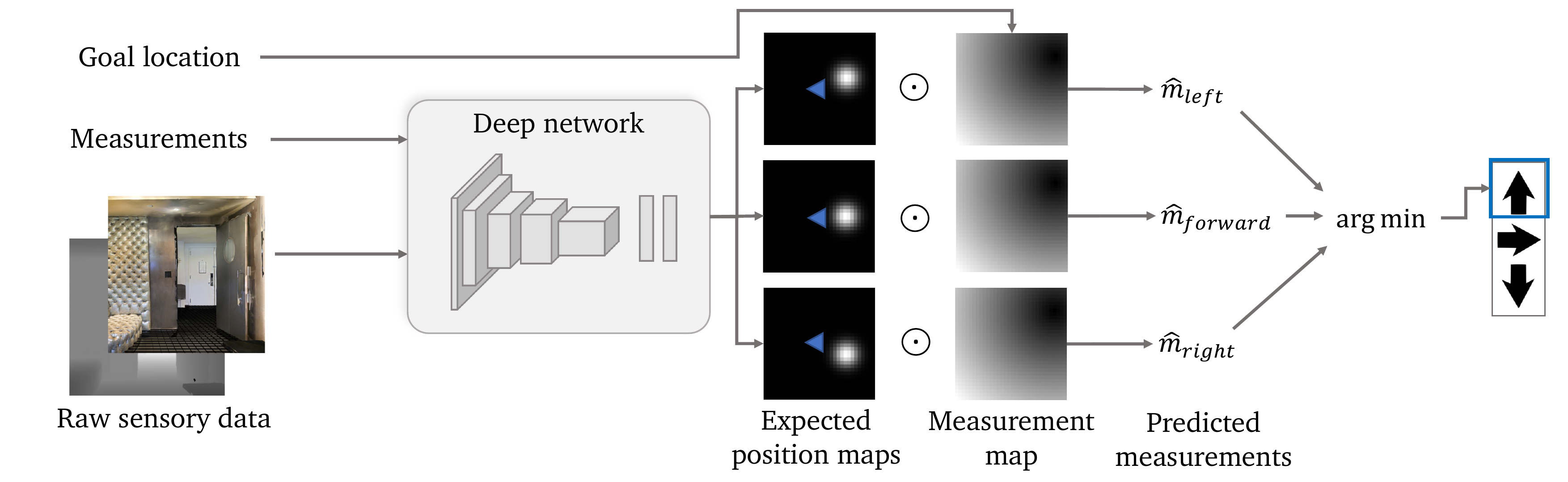}
    \vspace{2mm}
	\caption{A schematic visualization of \bdfpfull{}. Instead of directly regressing future measurements, we predict maps of the expected future location of the agent. By computing the dot product of these belief maps with the measurement maps, we get the predicted measurement values. In practice, we use measurement maps that encode the Euclidean distance to the goal.}
	\vspace{-2mm}
	\label{fig:attention_dfp}
\end{figure*}

\begin{figure}
	\centering
        \includegraphics[width=0.9\linewidth]{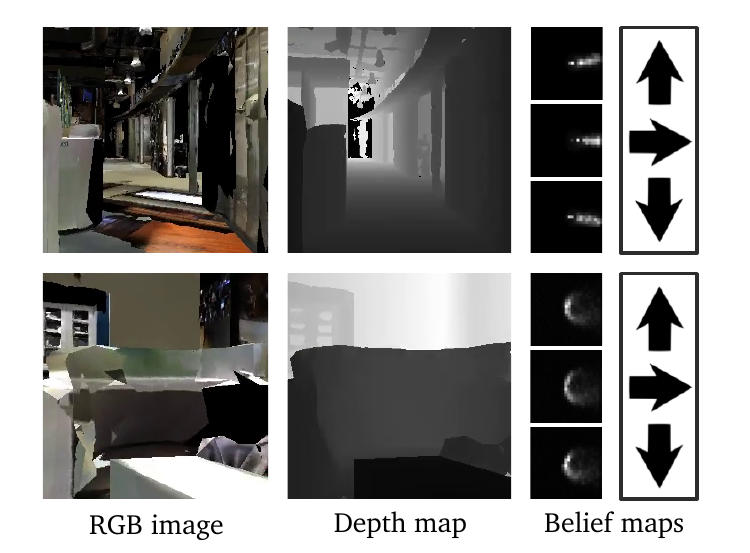}
	\caption{Examples of action-conditional expected position maps predicted by the \bdfp{} agent on Matterport3D. When facing free space (top), the agent expects to move forward, turning left or right depending on the action. When facing an obstacle (bottom), the agents expect to go around it on the left or on the right, but fails to predict different consequences of different actions.}
	\vspace{-2mm}
	\label{fig:belief}
\end{figure}

A schematic visualization of \bdfp{} is shown in Figure~\ref{fig:attention_dfp} and its operation is illustrated in Figure~\ref{fig:belief}.
Future measurement prediction is decomposed into two steps.
First, for each available action $\aa \in \aA$ the agent predicts belief maps $\BB^\aa_{t,\tau}$ of the expected agent's position after $\tau$ time steps when taking action $\aa$ at the current time step $t$.
We then compute the dot product of the predicted belief maps with the measurement map $\MM_t$ containing measurement values associated with each position: $\hat{\mm}_{t+\tau}^\aa = \BB^\aa_{t,\tau} \cdot \MM_t$.
These dot products are used as future measurement predictions in DFP.
Both maps are represented in the agent's egocentric coordinate system, with the agent located in the center of the map, facing right.
For the navigation task, we set the measurement maps $\MM_t$ to store the Euclidean distance to the goal, which can be trivially computed given the location of the goal relative to the agent.
Therefore, the dot product equals the expected distance to the goal.
We select the action that minimizes this expected distance.

\bdfp{} can be trained in the same way as standard DFP~-- by imposing a loss on the prediction of future measurements.
However, we can additionally train the expected future position maps directly, assuming we have access to the corresponding ground truth at training time.
We use both these losses in our experiments.
Further details are provided in the supplement.

\section{Implementation Details}
\label{sec:implementation_supp}
\subsection{Classic agent}
\mypara{Localization.} ORBSLAM2 and mapper parameters are shown in Table~\ref{tbl:orbslam_params}.

Because of monocular SLAM systems inherently output scale-less point cloud and the slam-to-world scale is typically set by detecting some kind of calibration pattern.  Since the environment is unknown and there is no such pattern, we designed the following procedure for RGB-only case.  After SLAM initialization, ``look down'' action is executed until the camera looks vertically down.  Median y-coordinate of observed keypoints is linked to the (known) agent height, providing needed scale.  Then the ``look up'' action is executed until horizontal orientation is recovered.  This procedure takes some time from the total time budget of the episode (approximately 4 seconds in our experiments).  In case of SLAM reset, the procedure needs to be done again.

\mypara{Locomotion.} Waypoint on the planned path at a fixed distance $d_1 = 0.5$ meters from the agent is selected. If the agent's orientation is not within $\phi = 15^{\circ}$ from the direction to the waypoint, the agent rotates towards the waypoint. Otherwise, the agent goes forward. If the distance to the waypoint is below  $d_2 = 0.15$ meters, a new waypoint is selected. If planned path changes, the waypoint position is updated accordingly.
Finally, with probability $p = 0.1$ random action is selected.

\begin{table}
\centering
{\small
\begin{tabular}{lccc}
\toprule
&& RGB-D & RGB \\
\midrule
Resolution && 256x256 & 512x512 \\
Max. \#features && 1000 & 5000\\
Min.FAST threshold && 1 & 1\\
Obstacle $H_{\text{min}}$ [m] && 0.1 &  0.1\\
Obstacle $H_{\text{max}}$ [m] &&  1.145 &  1.145\\
Points to create obstacle && 128 & 30 \\
\bottomrule
\end{tabular}
}
\vspace{2mm}
\caption{ORBSLAM2 agent parameters used in experiment.}
\vspace{-2mm}
\label{tbl:orbslam_params}
\end{table}

\mypara{Monocular depth estimation.} We use the ResNet-50~\cite{He2015} of Hu et al.~\cite{Hu2019}, pre-trained on the NYU dataset (provided by the authors) and then fine-tuned on data extracted from MINOS, separately on SunCG and Matterport3D. Input image size is 320x320 pixels, output is then resized to 256x256. We set to zeros all depth estimation, which are less than 0.1 meter and more than 3 meters. The estimated depth map is provided to both ORBSLAM2 and mapper as an input.

\begin{figure}
   \setlength{\tabcolsep}{0.5mm}
 	{\small
   \centering
   \begin{tabular}{ccccc}
       & Input RGB & Mono & Stereo & Ground truth \\
       \rotatebox[origin=c]{90}{SunCG} &
       \raisebox{-.5\height}{\includegraphics[height=0.22\linewidth]{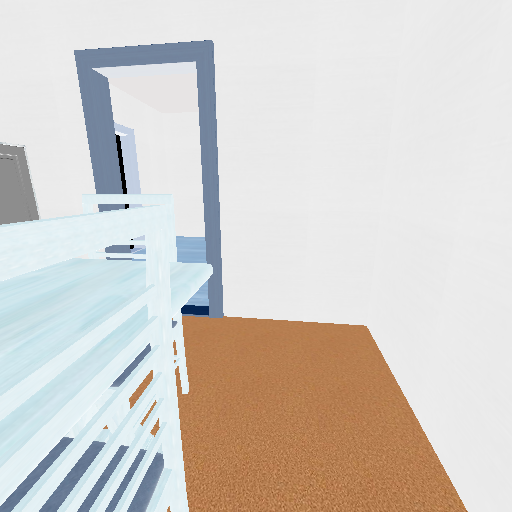}} &
       \raisebox{-.5\height}{\includegraphics[height=0.22\linewidth]{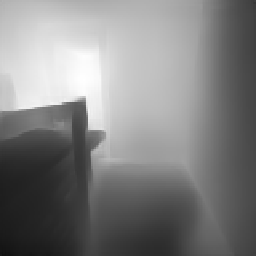}} &
       \raisebox{-.5\height}{\includegraphics[height=0.22\linewidth]{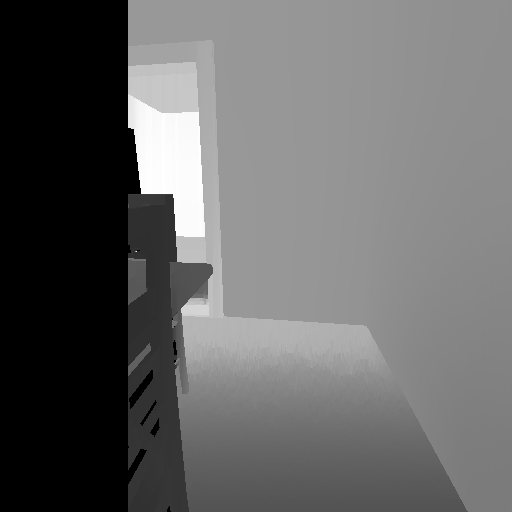}} &
       \raisebox{-.5\height}{\includegraphics[height=0.22\linewidth]{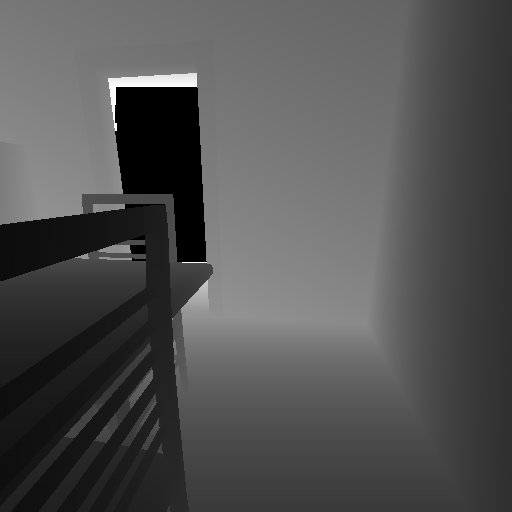}} \vspace*{1mm}\\
       \rotatebox[origin=c]{90}{M3D} &
       \raisebox{-.5\height}{\includegraphics[height=0.22\linewidth]{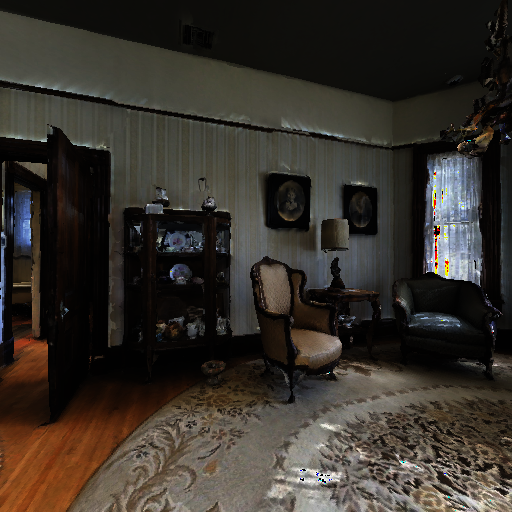}} &
       \raisebox{-.5\height}{\includegraphics[height=0.22\linewidth]{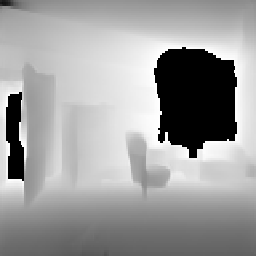}} &
       \raisebox{-.5\height}{\includegraphics[height=0.22\linewidth]{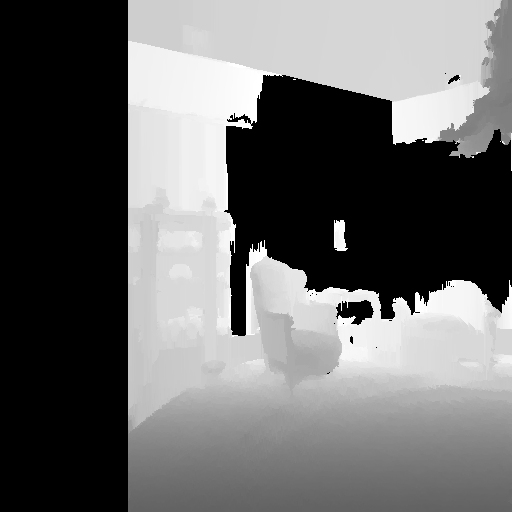}} &
       \raisebox{-.5\height}{\includegraphics[height=0.22\linewidth]{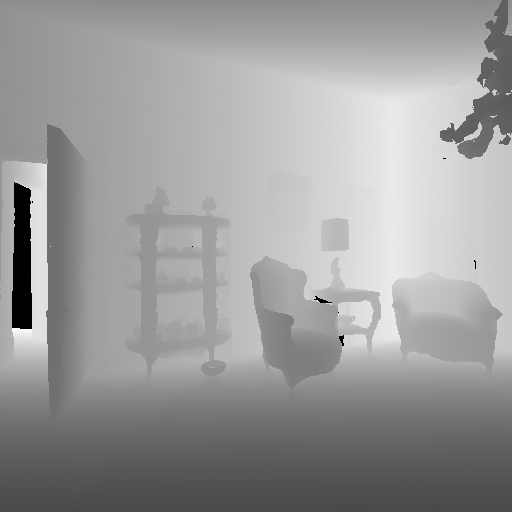}} \\
   \end{tabular}
   }
   \vspace{2mm}
   \caption{Typical estimated depth maps. Textureless regions and repetitive textures can lead to major errors when using stereo matching (see for instance wall n the top row). Even if the global structure is correct, metric imprecisions in depth estimation lead to drift of the SLAM system.}
   \label{fig:depth_estimation_qualitative}
 \end{figure}

\mypara{Stereo depth estimation.} 
We use OpenCV~\cite{OpenCV2000} StereoSGBM with weighted least squares disparity filter. The parameters are provided in Table~\ref{tbl:stereo}. The qualitive example of the estimated depth maps are shown in Figure~\ref{fig:depth_estimation_qualitative}.

\begin{table}
\centering
{\small
\begin{tabular}{lcc}
\toprule
Parameter && Value  \\
\midrule
Resolution && 512x512 \\
minDisparity && 0\\
numDisparities && 64\\
windowSize && 5\\
P1 && $24\times25$\\
P2 && $96\times25$\\
uniquenessRatio && 15\\
speckleWindowSize && 0\\
speckleRange && 2\\
preFilterCap && 64\\
mode && STEREO-SGBM-MODE-SGBM-3WAY\\
lambda && 80000\\
sigma && 1.2\\
visualMultiplier && 1.0\\
\bottomrule
\end{tabular}
}
\vspace{2mm}
\caption{Stereo depth estimation StereoSGBM weighted least squares disparity filter parameters used in experiment.}
\vspace{-2mm}
\label{tbl:stereo}
\end{table}

\subsection{Learned agent}
Architectures of the feature extraction and action-expectation networks are shown in Tables~\ref{tbl:DFP-feature-arch},\ref{tbl:action-expectation-networks}, \ref{tbl:action-expectation-belief-networks}. We observed no benefit in color images compared to grayscale, so grayscale is used for all learned agents. All modalities are composed by stacking current and 3 previous inputs. We found that Group Normalization~\cite{WuHe2018} makes training more stable and less sensitive to hyperparameters. In case of \bdfpfull{}, each 16x16 belief map is normalized by softmax function.


Loss is MSE loss on measurements: distance to the goal, sin and cos of the direction to the goal. In case of \bdfpfull{} additional MSE loss on belief map is applied.

\begin{table}
  \centering
  \small
  \ra{1.1}
    \begin{tabular}{@{}l@{\hspace{8mm}}cccc@{}}
      \toprule
     \multicolumn{4}{c}{Visual and depth networks, input = 128x128x4}\\
      \midrule
  \multicolumn{4}{c}{Conv2d 9x9, 32 channels, stride = 4, pad = 0}\\
       \multicolumn{4}{c}{LReLU(0.3), GN (4 groups)} \\
 \multicolumn{4}{c}{Conv2d 3x3, 64 channels, stride = 2, pad = 1}\\
       \multicolumn{4}{c}{LReLU(0.3), GN (8 groups)} \\
        \multicolumn{4}{c}{Conv2d 3x3, 128 channels, stride = 2, pad = 1}\\
       \multicolumn{4}{c}{LReLU(0.3), GN (8 groups)} \\
        \multicolumn{4}{c}{Conv2d 3x3, 64 channels, stride = 1, pad = 1}\\
       \multicolumn{4}{c}{LReLU(0.3), GN (8 groups)} \\
        \multicolumn{4}{c}{Conv2d 8x8, 256 channels, stride = 1, pad = 0}\\
        \midrule
          \multicolumn{4}{c}{Action history and measurements network, input = 4*3}\\
        \midrule
   \multicolumn{4}{c}{Fully connected, 128} \\
   \multicolumn{4}{c}{LReLU(0.3), GN (8 groups)} \\
   \multicolumn{4}{c}{Fully connected, 128} \\
   \multicolumn{4}{c}{LReLU(0.3), GN (8 groups)} \\
   \multicolumn{4}{c}{Fully connected, 128} \\
        \midrule
      \multicolumn{4}{c}{Goal network, input = 3}\\
        \midrule
   \multicolumn{4}{c}{Fully connected, 128} \\
   \multicolumn{4}{c}{LReLU(0.3), GN (8 groups)} \\
   \multicolumn{4}{c}{Fully connected, 128} \\
   \multicolumn{4}{c}{LReLU(0.3), GN (8 groups)} \\
   \multicolumn{4}{c}{Fully connected, 128} \\
   \bottomrule
    \end{tabular}
    \vspace{3mm}
  \caption{Feature networks architecture. GN stands for GroupNorm~\cite{WuHe2018}, LReLU(0.3) for max($x$, $0.3x$)~\cite{Xu2015}.}
  \label{tbl:DFP-feature-arch}
\end{table}

\begin{table}
  \centering
  \small
  \ra{1.1}
    \begin{tabular}{@{}l@{\hspace{8mm}}cccc@{}}
      \toprule
     \multicolumn{4}{c}{ExpectationNet }\\
      \midrule
   \multicolumn{4}{c}{Fully connected, 256} \\
   \multicolumn{4}{c}{LReLU(0.3), GN (16 groups)} \\
   \multicolumn{4}{c}{Fully connected, 256} \\
   \multicolumn{4}{c}{LReLU(0.3), GN (16 groups)} \\
   \multicolumn{4}{c}{Fully connected, 6*3} \\
        \midrule
     \multicolumn{4}{c}{ActionNet }\\
      \midrule
   \multicolumn{4}{c}{Fully connected, 256} \\
   \multicolumn{4}{c}{LReLU(0.3), GN (16 groups)} \\
   \multicolumn{4}{c}{Fully connected, 256} \\
   \multicolumn{4}{c}{LReLU(0.3), GN (16 groups)} \\
   \multicolumn{4}{c}{Fully connected, 6*6*3} \\

   \bottomrule
    \end{tabular}
    \vspace{3mm}
  \caption{Action-expectation networks architecture for plain DFP agents.   GN stands for GroupNorm~\cite{WuHe2018}, LReLU(0.3) for max($x$, $0.3x$)~\cite{Xu2015}.}
  \label{tbl:action-expectation-networks}
\end{table}

\begin{table}
  \small
  \centering
  \ra{1.1}
    \begin{tabular}{@{}l@{\hspace{8mm}}cccc@{}}
      \toprule
         \multicolumn{4}{c}{ExpectationBeliefNet }\\
      \midrule
   \multicolumn{4}{c}{Fully connected, 256} \\
   \multicolumn{4}{c}{LReLU(0.3), GN (16 groups)} \\
   \multicolumn{4}{c}{Fully connected, 256} \\
   \multicolumn{4}{c}{LReLU(0.3), GN (16 groups)} \\
   \multicolumn{4}{c}{Fully connected, 6*16*16} \\
        \midrule
     \multicolumn{4}{c}{ActionBeliefNet }\\
      \midrule
   \multicolumn{4}{c}{Fully connected, 256} \\
   \multicolumn{4}{c}{LReLU(0.3), GN (16 groups)} \\
   \multicolumn{4}{c}{Fully connected, 256} \\
   \multicolumn{4}{c}{LReLU(0.3), GN (16 groups)} \\
   \multicolumn{4}{c}{Fully connected, 6*6*16*16} \\
   \bottomrule
    \end{tabular}
    \vspace{3mm}
  \caption{Action-expectation networks architecture for plain DFP agents.   GN stands for GroupNorm~\cite{WuHe2018}, LReLU(0.3) for max($x$, $0.3x$)~\cite{Xu2015}.}
  \label{tbl:action-expectation-belief-networks}
\end{table}

\subsection{Experimental setup}
\begin{figure*}
	\centering
	\begin{tabular}{ccc}
	SunCG Empty & SunCG Furnished & Matterport3D \\
    \includegraphics[width=0.3\linewidth]{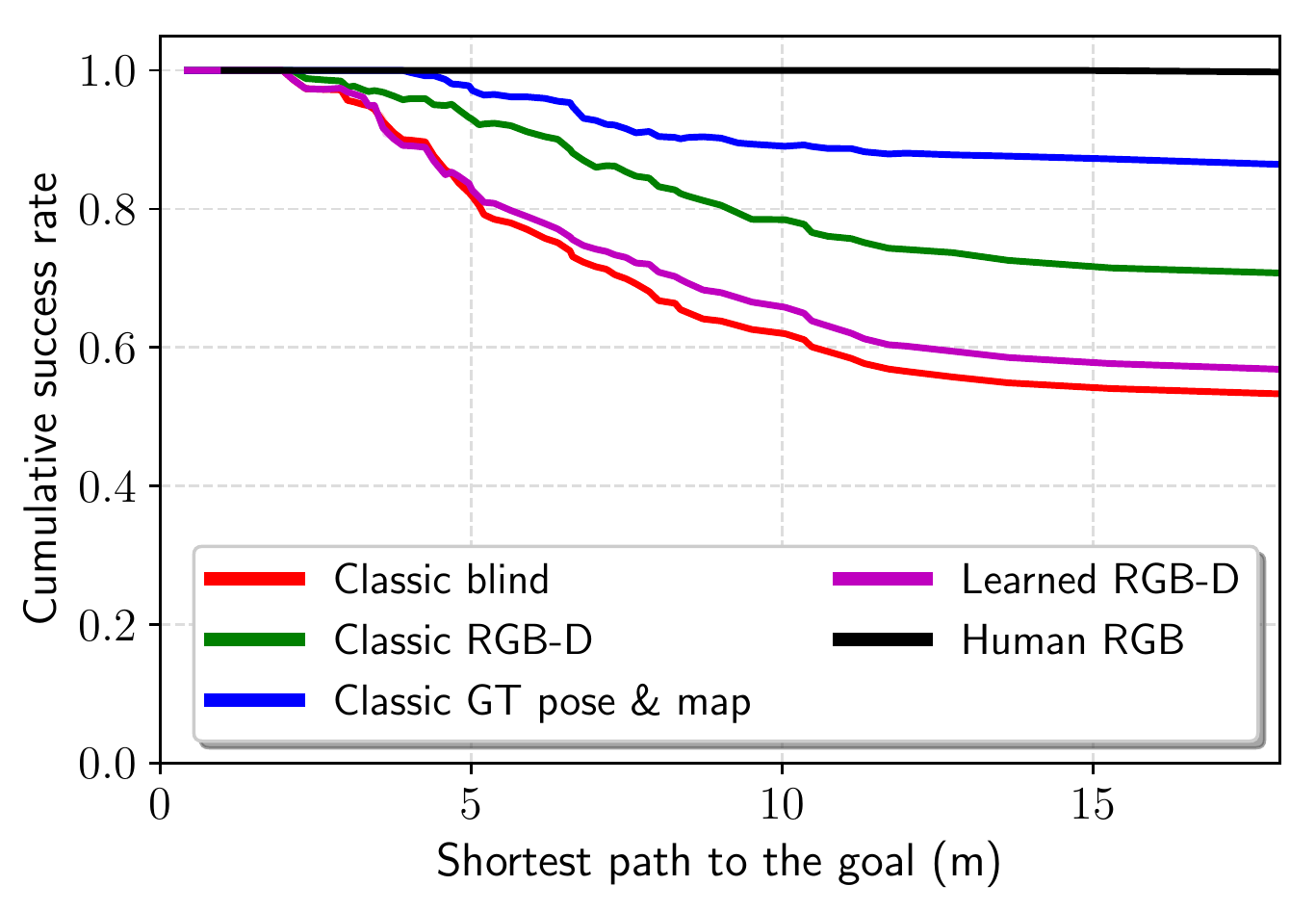} &   \includegraphics[width=0.3\linewidth]{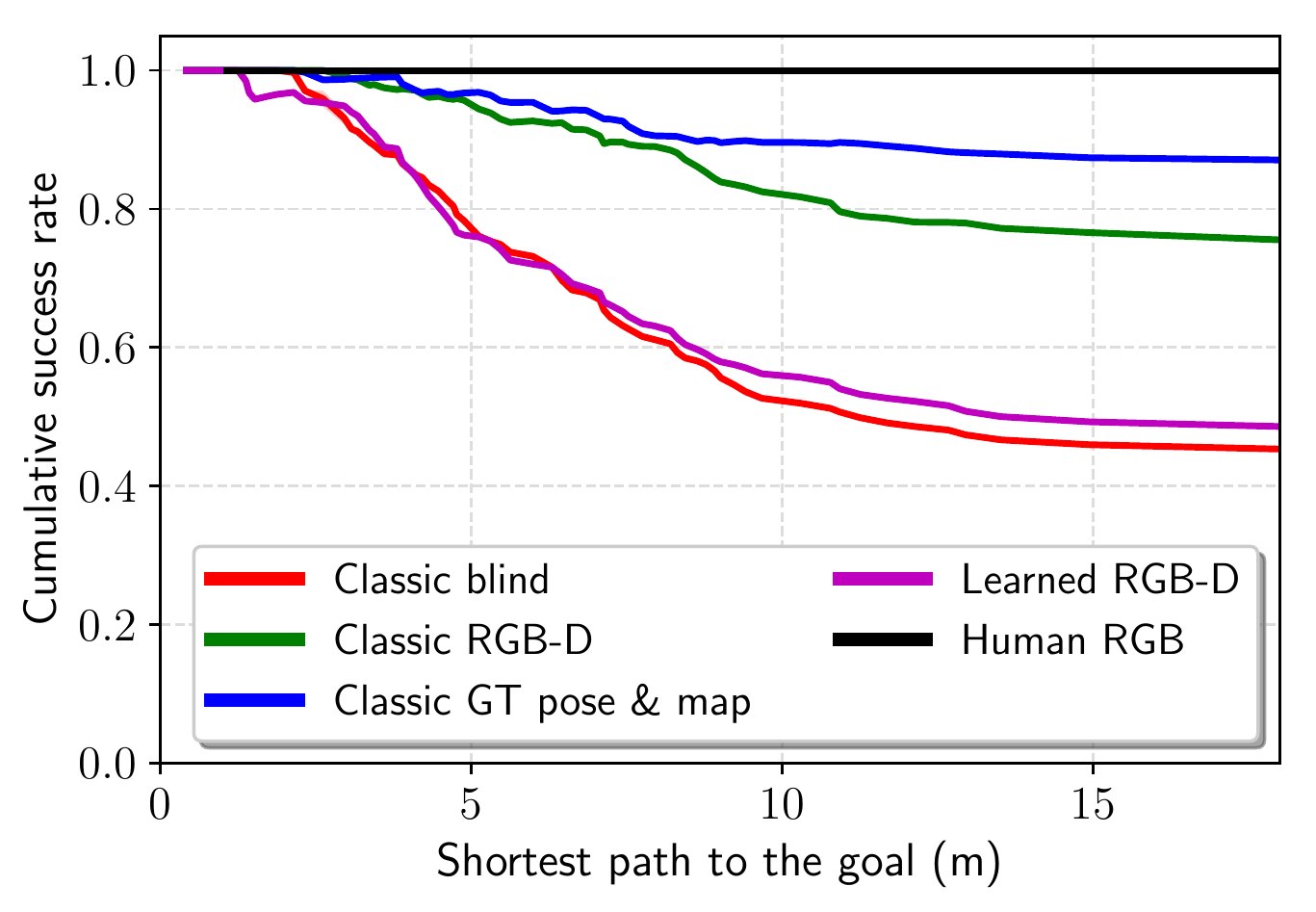} &
    \includegraphics[width=0.3\linewidth]{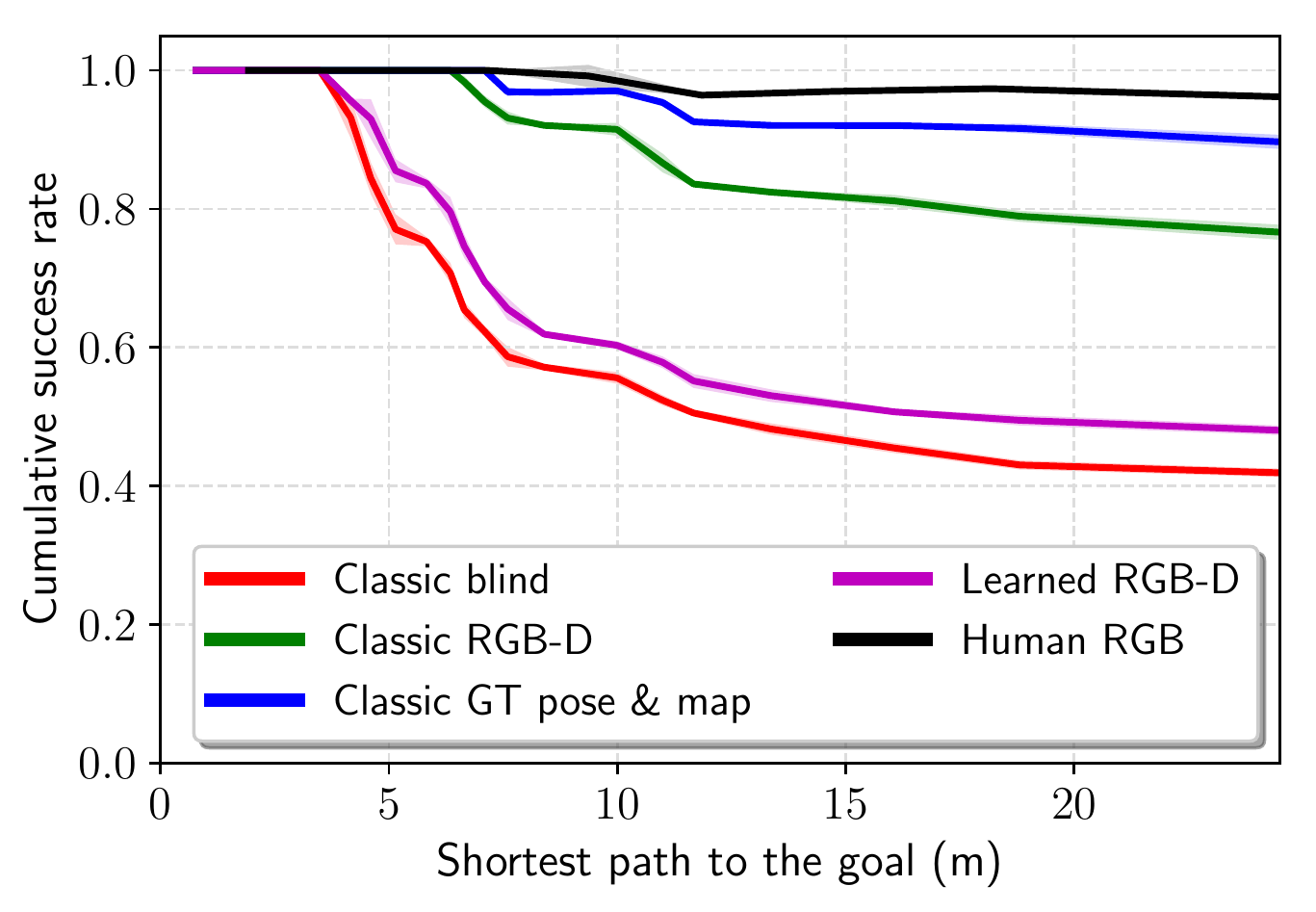}\\
    \includegraphics[width=0.3\linewidth]{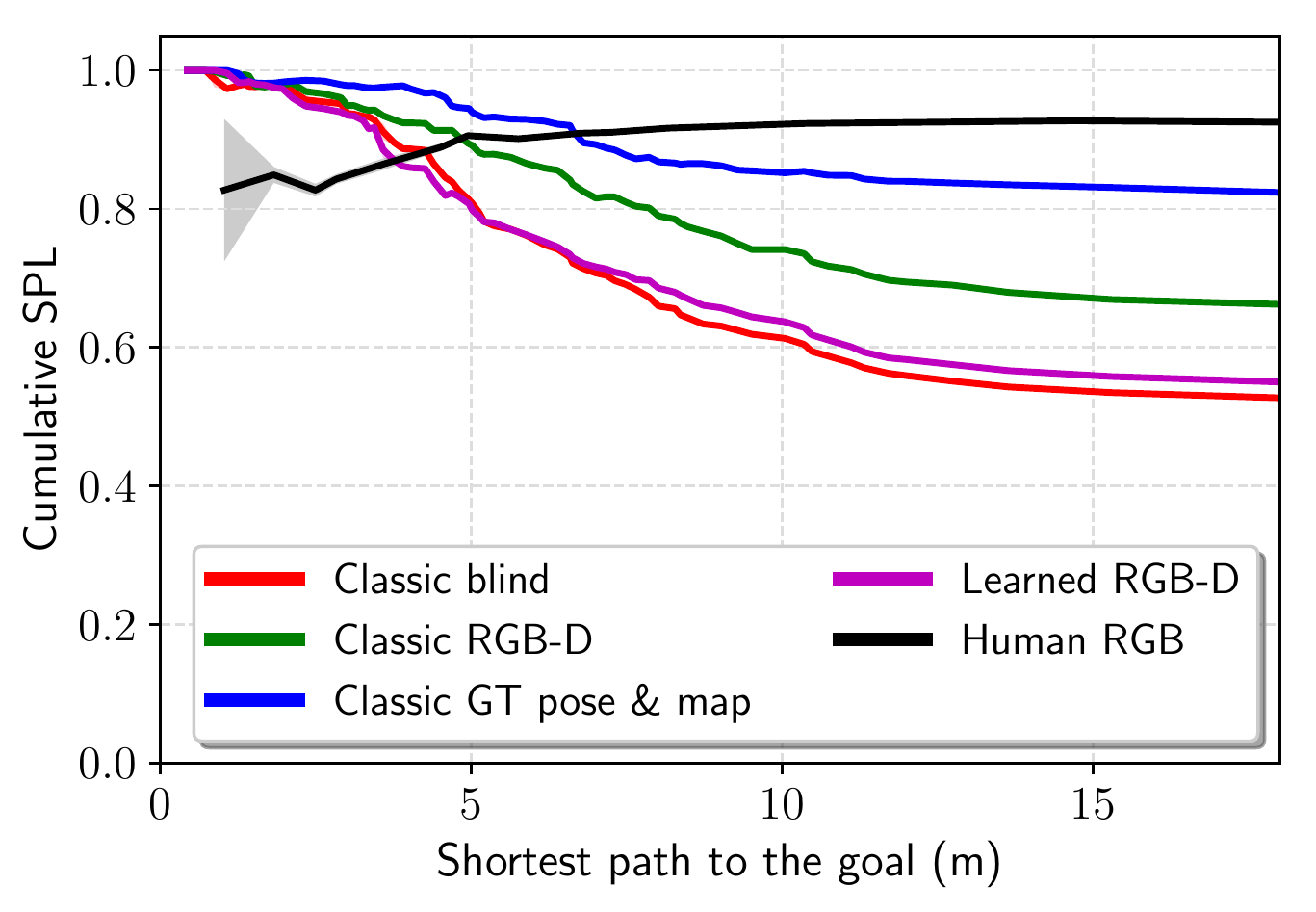} &
    \includegraphics[width=0.3\linewidth]{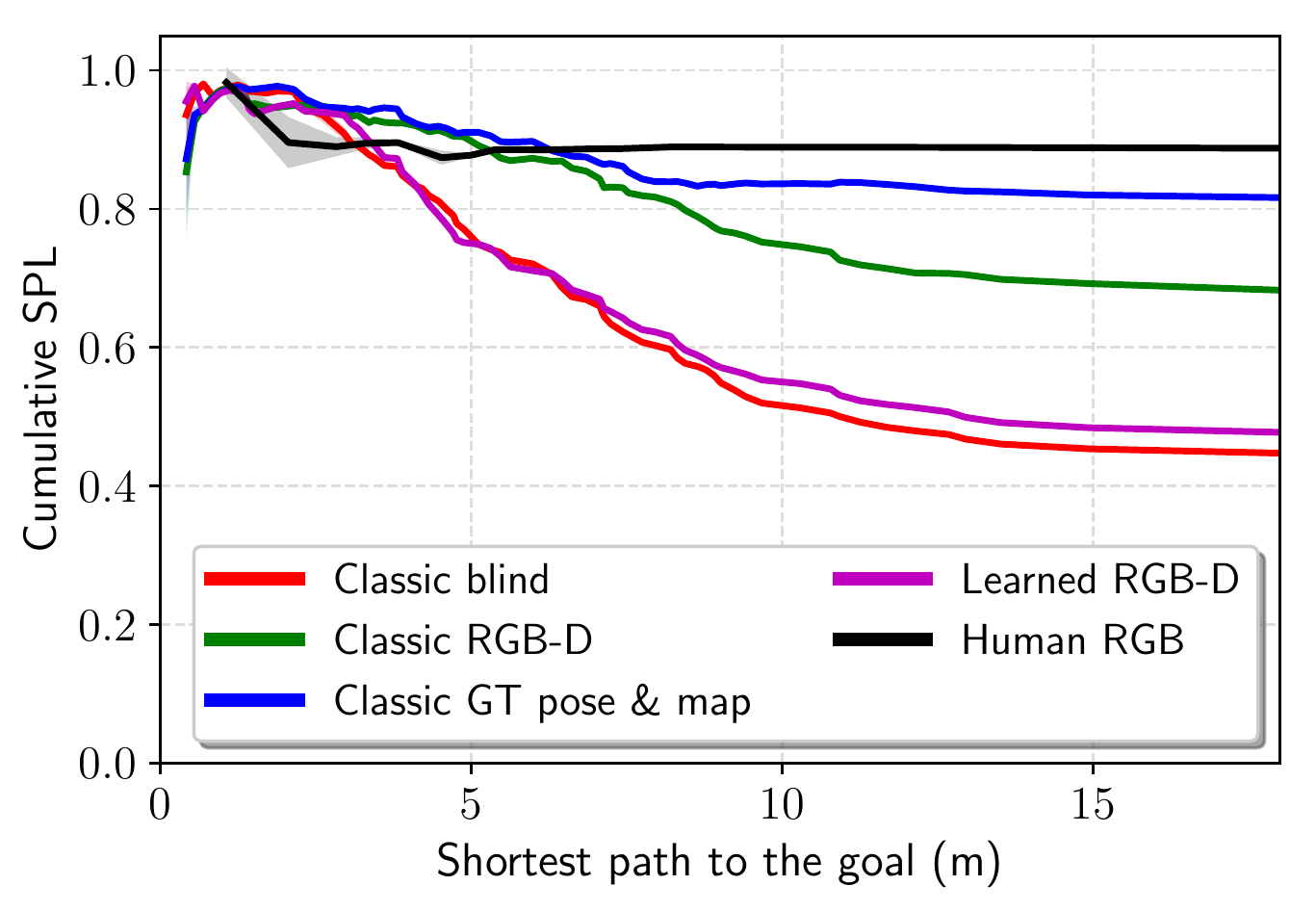} &
    \includegraphics[width=0.3\linewidth]{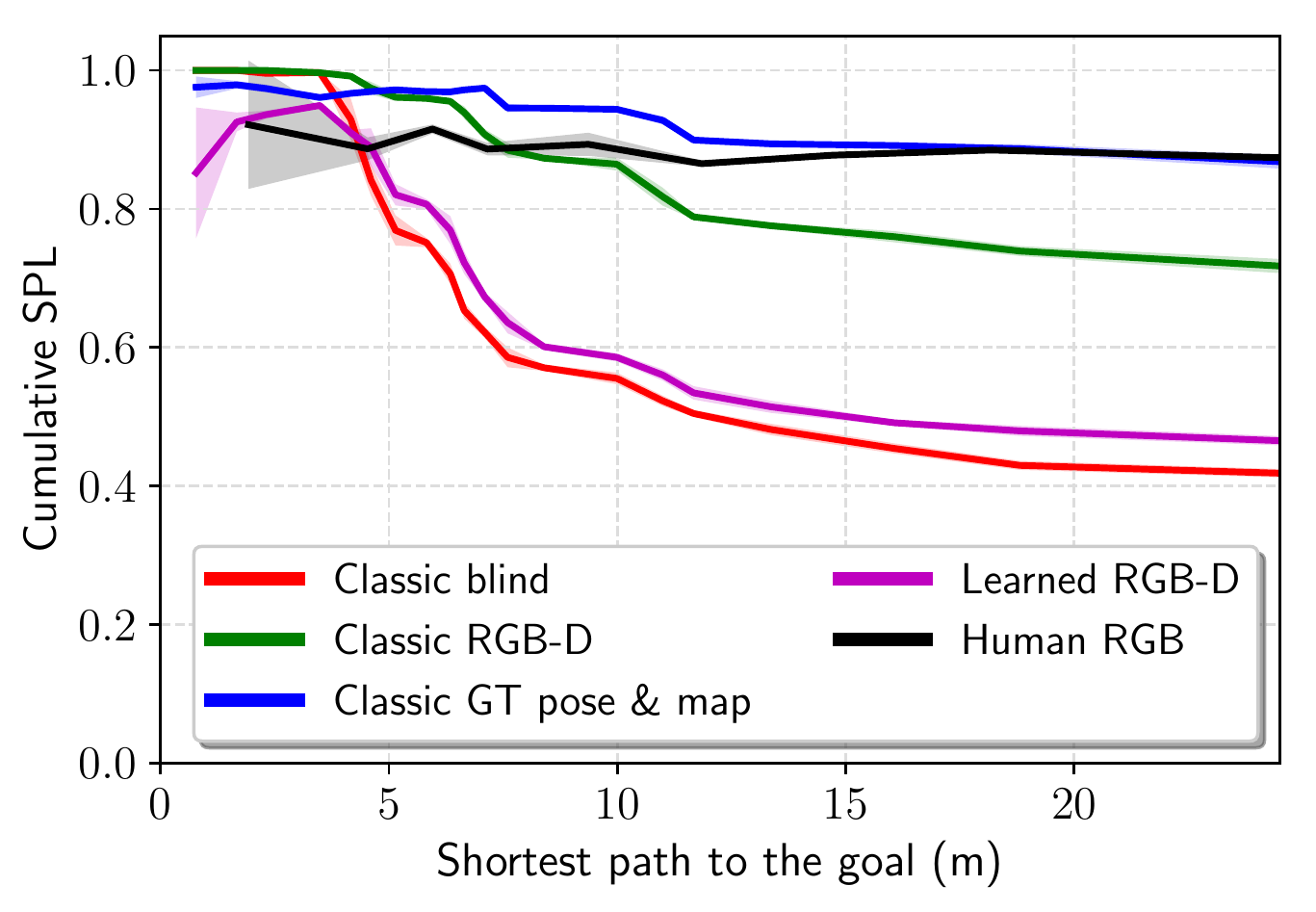}\\
    \includegraphics[width=0.3\linewidth]{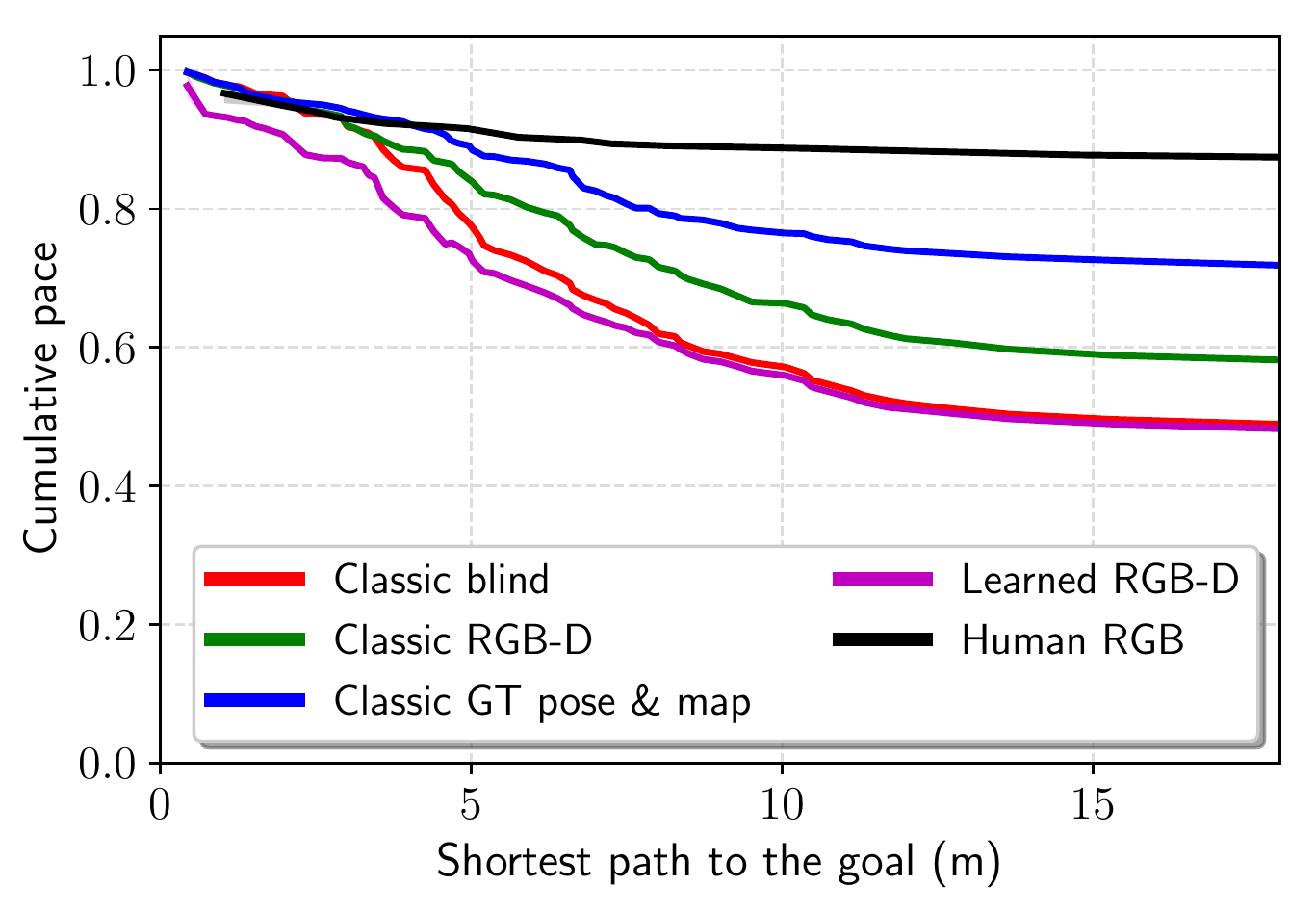} &
    \includegraphics[width=0.3\linewidth]{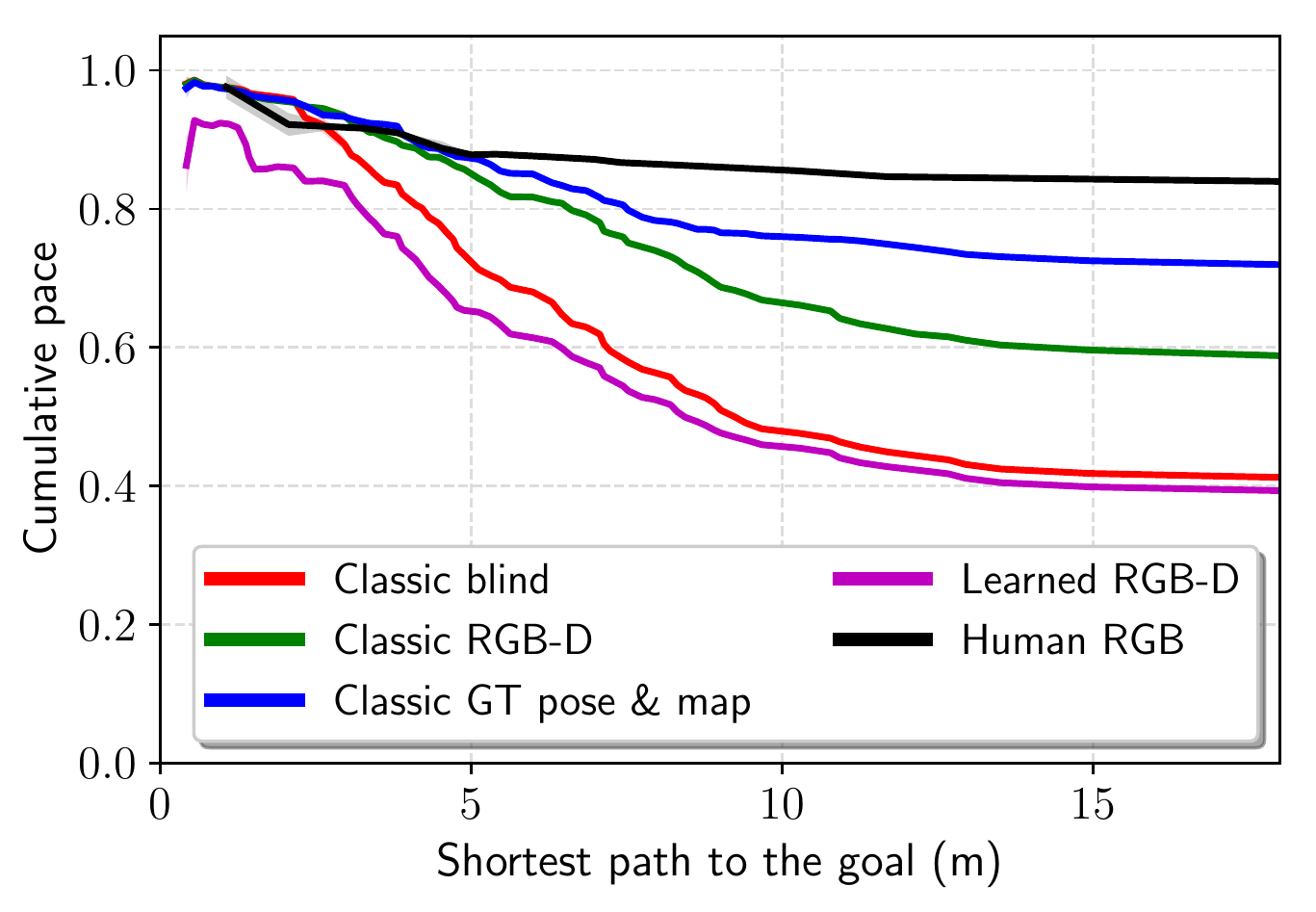} &
    \includegraphics[width=0.3\linewidth]{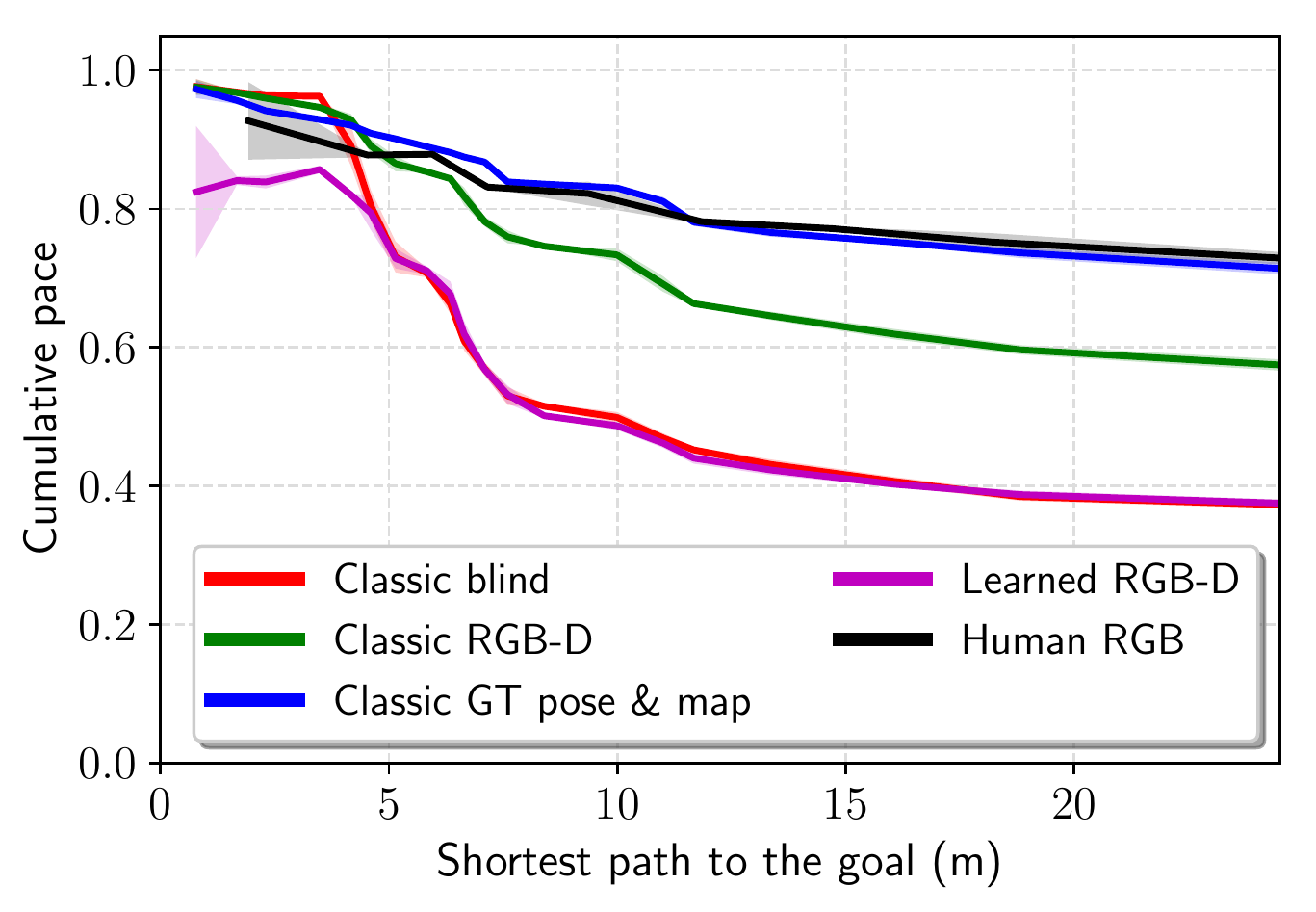}\\
    \end{tabular}

	\caption{Performance of algorithms on SunCG Empty (left), SunCG Furnished (middle) and Matterport3D (right) as a function of the length of the shortest path to the goal. We plot cumulative success rate (top), cumulative SPL (middle) and cumulative pace (bottom).}
  \vspace{-3mm}
	\label{fig:cumulative}
\end{figure*}

\paragraph{Metrics.}
We use three main evaluation metrics in our experiments: success rate, success weighted by path length (SPL)~\cite{Anderson2018}, and ``pace'' measuring how fast the agent reaches the goal.
Value of each of the metrics is between 0 and 1, and in what follows we report all metrics in \%.

For each of the $N$ evaluation trials denote by $S_i$ the binary success indicator, by $\ell_i$ the shortest path from start to the goal, by $p_i$ the length of the actual executed path, and by $T_i$.

\textbf{Success rate} is simply the fraction of episodes in which the agent manages to reach the goal within the given time budget:
\begin{equation}
    \text{SR} = \frac{1}{N} \sum_{i=1}^N S_i
\end{equation}

\textbf{SPL} takes into account the length of the executed trajectories, not just the binary success indicator:
\begin{equation}
  \text{SPL} =   \frac{1}{N} \sum_{i=1}^N S_i \frac{\ell_i}{\max (\ell_i, p_i)}.
\end{equation}
This is a stricter metric: to achieve a perfect score of $1$, the agents needs not only reach the goal in every episode, but also follow the shortest path.

\textbf{Pace} is the average time left unused after the episode is finished, divided by available time budget (500 steps):
\begin{equation}
    \text{P} = \frac{1}{N}  \sum_{i=1}^N (1 - T_i).
\end{equation}
The higher this number, the faster, on average, the agent reaches the goal.
An important advantage of this metric over SPL is that it takes into account time spent standing still or rotating. A disadvantage is that the perfect score of $1$ is not achievable: it would correspond to instantly teleporting to the goal.

\section{Additional Results}
\label{sec:experiments_supp}

We show success rate and pace for different modalities in Tables~\ref{tbl:sr_results_modalities},\ref{tbl:pace_results_modalities}.
All metrics for all environments as a function of the shortest path to the goal are shown in Figure~\ref{fig:cumulative}.
Note that on SPL the human performance is suboptimal for short path lengths.
This is because 1) the human often takes some steps in the beginning of an episode to self-localize, and 2) the goal representation in polar coordinates is not very intuitive, in particular, estimating distances is difficult, which sometimes leads to suboptimal trajectories.
However, for longer trajectories the robustness of human navigation compensates for these drawbacks.

\begin{table*}
\centering
{\small
\begin{tabular}{lcccccccccccc}
\toprule
& & \multicolumn{2}{c}{SunCG Empty} & & \multicolumn{2}{c}{SunCG Furnished} & & \multicolumn{2}{c}{Matterport3D} & & \multicolumn{2}{c}{Mean} \\
& &  Classic & Learned & & Classic & Learned & & Classic & Learned & & Classic & Learned \\ \midrule
Blind  &&  52.9& \textbf{58.2} &&  \textbf{44.9} & 16.4 && \textbf{40.9} & 30.0 && \textbf{46.3} & 34.9   \\
RGB    &&  9.4 & \textbf{49.9} &&  10.0 & \textbf{44.8} &&9.0 & \textbf{37.4} && 9.4 & \textbf{44.0}   \\
RGB-D  &&  \textbf{70.2} & 56.4 &&  \textbf{75.0} & 48.2 && \textbf{75.0} & 46.9 && \textbf{73.4} & 50.5  \\
\midrule
RGB-D + GT pose        && 90.8 & -- && 92.0 & -- && 87.9 & -- && 90.3 & --  \\
RGB-D + GT pose \& map && 85.8 & -- && 87.0 & -- && 87.9 & -- && 86.9 & --   \\
\bottomrule
\end{tabular}
}
\vspace{2mm}
\caption{Average success rate. Performance of a classic modular pipeline and a learned agent when provided with different input modalities. For each environment and sensory input, the best of the two results is highlighted in bold. }
\vspace{-2mm}
\label{tbl:sr_results_modalities}
\end{table*}
\begin{table*}
\centering
{\small
\begin{tabular}{lcccccccccccc}
\toprule
& & \multicolumn{2}{c}{SunCG Empty} & & \multicolumn{2}{c}{SunCG Furnished} & & \multicolumn{2}{c}{Matterport3D} & & \multicolumn{2}{c}{Mean} \\
& &  Classic & Learned & & Classic & Learned & & Classic & Learned & & Classic & Learned \\ \midrule
Blind  &&  \textbf{48.5} & 46.7 &&  \textbf{40.9} & 9.5 && \textbf{36.4} & 22.7 && \textbf{41.9} & 26.3   \\
RGB    &&  7.2 & \textbf{49.9} && 8.2 & \textbf{44.8} && 8.3 & \textbf{37.4} && 7.9 & \textbf{44.0}   \\
RGB-D  &&  \textbf{57.8} & 47.9 &&  \textbf{58.3} & 39.0 && \textbf{56.2} & 36.6 && \textbf{57.4} & 41.2  \\
\midrule
RGB-D + GT pose        && 71.9 & -- && 72.1 & -- && 64.7 & -- && 69.5 & --  \\
RGB-D + GT pose \& map && 71.4 & -- && 71.6 & -- && 70.0 & -- && 71.0 & --   \\
\bottomrule
\end{tabular}
}
\vspace{2mm}
\caption{Average pace. Performance of a classic modular pipeline and a learned agent when provided with different input modalities. For each environment and sensory input, the best of the two results is highlighted in bold. }
\vspace{-2mm}
\label{tbl:pace_results_modalities}
\end{table*}



\end{document}